\documentclass{article}

\usepackage[nonatbib, main, final]{neurips_2026}
\usepackage{amsmath}

\usepackage[utf8]{inputenc} %
\usepackage[T1]{fontenc}    %
\usepackage{hyperref}       %
\usepackage{url}            %
\usepackage{booktabs}       %
\usepackage{amsfonts}       %
\usepackage{nicefrac}       %
\usepackage{microtype}      %
\usepackage{xcolor}         %

\usepackage{microtype}
\usepackage{graphicx}
\usepackage{subcaption}

\usepackage{amssymb}
\usepackage{mathtools}
\usepackage{amsthm}
\usepackage{algorithm}
\usepackage{algpseudocode}
\usepackage{cancel}
\usepackage{multirow}
\usepackage{enumitem}

\usepackage[capitalize,noabbrev]{cleveref}

\theoremstyle{plain}
\newtheorem{theorem}{Theorem}[section]

\theoremstyle{definition}

\theoremstyle{remark}

\usepackage{wrapfig}
\usepackage{caption}   %
\usepackage{booktabs}
\usepackage{tabularx}
\usepackage{array}

\title{\textsc{DiffRatio}: Training One-Step Diffusion Models Without Teacher Supervision}

\author{%
  Wenlin Chen\thanks{Equal contribution. Correspondence to Wenlin Chen \texttt{<wenlin.chen@cn.bosch.com>}, Mingtian Zhang \texttt{<m.zhang@cs.ucl.ac.uk>}, and Jiajun He \texttt{<jh2383@cam.ac.uk>}.} \\
  Bosch Center for AI\\
  \And
  Mingtian Zhang$^*$ \\
  University College London \\
  \And
  Jiajun He$^*$ \\
  University of Cambridge \\
  \AND
  Zijing Ou \\
  Imperial College London \\
  \And
  José Miguel Hernández-Lobato \\
  University of Cambridge \\
  \AND
  Bernhard Schölkopf \\ 
  Max Planck Institute for Intelligent Systems \\
  \And
  David Barber \\
  University College London
}

\begin{document}

\maketitle

\begin{abstract}
Score-based distillation methods train one-step diffusion models in two stages: they first train a teacher score model, then distill it into a one-step student model. However, this distillation process can introduce bias from two sources: errors in the teacher score model and student score estimate.
We propose \textbf{\textsc{DiffRatio}}, a new approach for training one-step diffusion models without teacher supervision based on a density-ratio formulation that connects score-based distribution matching and GAN-style adversarial training.
Instead of using a teacher score model to provide training targets, DiffRatio directly learns a log density ratio between the student and data distributions across diffusion time steps. As a pre-trained score model is only used to initialize the one-step generator and does not supervise training, this design simplifies the training pipeline, mitigates gradient estimation bias, and reduces the size of auxiliary networks.
The learned density ratio can be used as a verifier, enabling a principled inference-time parallel scaling method that further improves sample quality without external rewards or extra sequential computation. DiffRatio achieves strong one-step generation results on CIFAR-10 and ImageNet ($64{\times}64$ and $512{\times}512$), outperforming most teacher-supervised distillation methods.
\end{abstract}

\section{Introduction}

Diffusion models~\cite{sohl2015deep,ho2020denoising,song2019generative} have achieved remarkable success in modeling complex real-world data across a wide range of domains~\cite{rombach2022high,li2022diffusion,pooledreamfusion,ho2022video,liu2023audioldm}. Typically, diffusion models consist of two processes: a forward noising process, which gradually perturbs data into a known noise prior (e.g., Gaussian), and a reverse denoising process, which learns to invert the forward corruption process to generate realistic data samples from noise. Formally, the forward process is defined over $T$ time steps as a Markov chain of Gaussian transitions, while the reverse process is parameterized by neural networks that predict the denoising distribution given the noisy samples.
In classic diffusion models with Gaussian denoising distributions, generating high-quality data samples typically requires hundreds or thousands of sampling steps, resulting in significant sampling inefficiency due to the need for $T \gg 1$ NFEs (number of function evaluations)~\cite{ho2020denoising,nichol2021improved}. To address this weakness, various acceleration methods have been proposed to reduce NFEs during sampling. One class of such approaches leverages advanced numerical solvers for differential equations, enabling continuous-time approximations of the diffusion process~\cite{song2020denoising,liu2022pseudo,lu2022dpm}. Another line of work improves the flexibility of the posterior distribution in the denoising process, either by estimating a more accurate covariance for the Gaussian distribution~\cite{nichol2021improved,bao2022analytic,bao2022estimating,ou2024improving} or by adopting flexible non-Gaussian denoising distributions~\cite{debortoli2025distributionaldiffusionmodelsscoring,xiao2021tackling,yu2024hierarchical}. While these techniques can dramatically reduce NFEs from $\sim10^3$ to around 10--20, they still fall short of achieving high-quality generation within 5 steps.

Recently, distillation-based methods have emerged as a powerful direction for training diffusion models, enabling high-quality \emph{one-step} generation~\cite{zhou2024score}. These methods fall into two categories.
Trajectory-based distillation methods~\cite{salimans2022progressive,berthelot2023tract,song2023consistency,heek2024multistep,kim2023consistency,frans2024one,li2024bidirectional,geng2025mean,boffi2025flow} aim to approximate the full sampling trajectory by training a student model to amortize multiple intermediate steps. These methods are motivated by accelerated solvers and typically perform joint training of the full diffusion model and the distillation process.
Score-based distillation approaches~\cite{luo2024diff,yin2024improved,salimans2024multistep,xie2024distillation,zhou2024score} distill the full denoising process of the pre-trained teacher diffusion model into a one-step latent variable model. This distillation process typically involves minimizing a divergence between the student and teacher models based on their respective score estimations~\cite{pooledreamfusion,wang2024prolificdreamer} and typically follow a two-stage pipeline: (1)~\emph{pre-training stage}: training a teacher score model via denoising score matching on the data distribution, and (2)~\emph{distillation stage}: using the pre-trained teacher score to supervise the training of a one-step student model. However, the distillation-stage gradient estimator suffers from two sources of bias: (1)~imperfect teacher score supervision due to score estimation error from pre-training, and (2)~student score estimation error during distillation. These biases can accumulate and thus significantly degrade the quality of the resulting one-step model. 

We propose \textbf{\textsc{DiffRatio}}, a new approach for training one-step diffusion models by directly estimating the \emph{score difference} between the student and data distributions through the gradient of a learned log-density-ratio network. As a pre-trained score model is only used to initialize the one-step generator and does not supervise training, this approach mitigates gradient estimation bias, simplifies the training pipeline, and reduces the size of auxiliary networks. The learned density ratio can further act as a verifier, enabling a \emph{principled inference-time parallel scaling scheme} that improves generation quality without external rewards or extra sequential computation. DiffRatio achieves state-of-the-art one-step generation performance across standard image generation benchmarks. 
Moreover, we find that initializing the student model with the teacher’s weights remains critical since the teacher diffusion model provides a rich set of feature representations across different noise levels. 

\section{Background}
\subsection{Diffusion Models}
Diffusion models~\cite{sohl2015deep, ho2020denoising, song2019generative} define a generative framework that maps a Gaussian prior $p(x_T)$ to the data distribution $p(x_0)$ through a forward noising process and a learned reverse denoising process. The forward process progressively adds Gaussian noise to the data
\begin{align}
    q(x_{0:T}) = p_d(x_0)\prod_{t=1}^T q(x_t | x_{t-1}),
\end{align}
with transition kernels defined as 
$
    q(x_t | x_{t-1}) = \mathcal{N}(x_t | \sqrt{1 - \beta_t} \, x_{t-1}, \beta_t I),
$
where \( \beta_t \in (0,1) \) is a pre-specified variance. The skip conditional distribution at time \( t \) can be written as:
$
    q(x_t | x_0) = \mathcal{N}(x_t | \sqrt{\bar{\alpha}_t} x_0, (1 - \bar{\alpha}_t) I),
$
where \( \bar{\alpha}_t = \prod_{s=1}^t (1 - \beta_s) \). As \( T \to \infty \), the final state \( x_T \) approximates a standard normal distribution, i.e., \( q(x_T) \to \mathcal{N}(0, I) \).
The generative process aims to reverse this trajectory. Starting from noise \( x_T \sim p(x_T) = \mathcal{N}(0, I) \), the model learns a reverse process to sequentially denoise and reconstruct data samples. Since the true reverse conditional \( q(x_{t-1} | x_t) \) is intractable, a common method is to approximate it with a variational Gaussian distribution:
\begin{align}
    p_\theta(x_{t-1} | x_t) = \mathcal{N}(x_{t-1} | \mu_{t-1}(x_t; \theta), \Sigma_{t-1}(x_t; \theta)),
\end{align}
where the mean function is learned from data and the covariance function can be either learned~\cite{nichol2021improved,ou2024improving,bao2022estimating} or chosen to be a fixed value~\cite{bao2022analytic,ho2020denoising}. 
From a score-based perspective, the denoising mean \(\mu_{t-1}(x_t;\theta)\) can be written in terms of the (time-dependent) score of the forward marginal \(q(x_t)\), i.e., \(\nabla_{x_t}\log q(x_t)\). In practice, we learn score networks using denoising score matching (DSM)~\cite{vincent2011connection, song2019generative}. Specifically, for any distribution \(q(x_0)\) and its noised marginal \(q(x_t)=\int q(x_t|x_0)\,q(x_0)\,dx_0\), DSM trains a score network \(s_\theta(x_t,t)\) to estimate \(\nabla_{x_t}\log q(x_t)\) by minimizing
\begin{align}
    \min_{\theta}~\mathbb{E}_{q(x_0)\,q(x_t|x_0)}\!\left[\left\|s_{\theta}(x_t,t) - \nabla_{x_t} \log q(x_t|x_0)\right\|_2^2\right].
\end{align}
Given an estimated score \(s_\theta(x_t,t)\approx \nabla_{x_t}\log q(x_t)\), Tweedie’s Lemma~\cite{efron2011tweedie, robbins1992empirical} yields the corresponding reverse mean:
\begin{align}
    \mu_{t-1}(x_t; \theta) = \frac{x_t + \beta_t\, s_\theta(x_t,t)}{\sqrt{1 - \beta_t}},
\end{align}
which establishes a connection between the denoising distributional perspective and the score estimation perspective of diffusion models. In the following section, we introduce score-based distillation methods through the lens of divergence minimization.

\subsection{Score-Based Diffusion Distillation}

Score-based distillation methods aim to distill a teacher diffusion model $p_\theta$ (pre-trained on the true data distribution $p_d(x_0)$) into a one-step implicit generative model~\cite{goodfellow2014generative,huszar2017variational,zhang2020spread}:
\begin{align}
    q_\theta(x_0) = \int \delta(x_0-g_\theta(z))p(z)dz,
\end{align}
where $\delta(\cdot)$ is the Dirac delta function, $p(z)$ is a standard Gaussian prior for the latent variable $z$, and $g_\theta:\mathcal{Z}\to\mathcal{X}$ is a deterministic neural network that generates data $x$ from the latent variable $z$ in one step. We want to minimize the divergence between this  one-step model and the true data distribution. However, when the function \( g_\theta(\cdot) \) is not bijective, the model distribution \( q_\theta \) is not absolutely continuous w.r.t. the Lebesgue measure. As a result, the corresponding density function may not be well-defined, and consequently, the commonly used $f$-divergence such as KL divergence between \( q_\theta(x_0) \) and the data distribution \( p_d(x_0) \) may also be ill-defined~\cite{arjovsky2017wasserstein,zhang2020spread}.

Inspired by diffusion models, one can use a set of (scaled) Gaussian convolution kernels $\mathcal{K}=\{k_1,\cdots, k_T\}$ with $k_t(x_t|x_0)=\mathcal{N}(x_t|\alpha_t x_0, \sigma_t^2I)$ to define the Diffusive KL divergence (DiKL) between the model density $q_\theta(x_0)$ and target distribution $p_d(x_0)$:
\begin{align}
    \mathrm{DiKL}_{\mathcal{K}}(q_\theta||p_d)
    \equiv\sum_{t=1}^T w(t)\mathrm{KL}( q_\theta(x_t)|| p_d(x_t)),
\end{align}
where $w(t)$ is a positive scalar weighting function that sums to one, and $q_\theta(x_t)$ and $p_d(x_t)$ are noisy model density and noisy target density, respectively, defined by
\begin{align}
    q_\theta(x_t)=\int k(x_t|x_0)q_\theta(x_0)dx_0, \quad
    p_d(x_t)=\int  k(x_t|x_0)p_d(x_0)dx_0.
\end{align}
In this case, the distributions $q_\theta(x_t)$ and $p_d(x_t)$ are always absolutely continuous, and thus the KL divergence between them is always well-defined. For a single Gaussian kernel, the divergence was known as \emph{Spread KL divergence}~\cite{zhang2020spread,zhang2019variational}. It is straightforward to show that it is a valid divergence, i.e., $\mathrm{DiKL}_{\mathcal{K}}(q_\theta||p_d)=0\Leftrightarrow q_\theta=p_d$; see \cite{zhang2020spread} for a proof. In addition to the diffusion distillation~\cite{luo2024diff, xie2024distillation,wang2025vardiu}, this divergence has successfully been used in 3D generative models~\cite{pooledreamfusion,wang2024prolificdreamer} and training neural samplers~\cite{he2024training}.
For a single kernel $k_t$, the DiKL term admits the following pathwise gradient (see Appendix~\ref{appendix:dikl} for a derivation):
\begin{align}
    \nabla_\theta \mathrm{DiKL}_{t}(q_\theta (x_0) || p_d(x_0))
    =\int q_\theta(x_t)\Big(\underbrace{\nabla_{x_t} \log q_\theta(x_t) - \nabla_{x_t} \log p_d(x_t)}_{\text{score difference at time }t}\Big) \frac{\partial x_t}{\partial \theta} dx_t.\label{eq:kl:gradient}
\end{align}
However, neither the noisy model score \( \nabla_{x_t} \log q_\theta(x_t) \) nor the noisy target score \( \nabla_{x_t} \log p_d(x_t) \) are directly accessible. In the distillation setting, the noisy target score is provided by a pre-trained diffusion model. Specifically, a score network \( s_{\psi_1}^{p_d}(x_t, t)\) is learned to approximate $\nabla_{x_t} \log p_d(x_t) $ by minimizing the DSM loss w.r.t. $\psi_1$ at time $t$:
\begin{align}
\min_{\psi_1}~\mathbb{E}_{p_d(x_0)k(x_t|x_0)}\left[\|s_{\psi_1}^{p_d}(x_t,t) - \nabla_{x_t} \log k(x_t|x_0)\|_2^2\right]. \label{eq:dsm_loss:pd}
\end{align}
Regarding the noisy model score $\nabla_{x_t} \log q_\theta(x_t)$, since samples from $q_\theta$ can be efficiently obtained, we can approximate the score with another score network $s_{\psi_2}^{q_\theta}(x_t,t) \approx \nabla_{x_t} \log q_\theta(x_t)$ by minimizing the DSM loss w.r.t. $\psi_2$:
\begin{align}
\min_{\psi_2}~\mathbb{E}_{q_\theta(x_0)k(x_t|x_0)}\left[\|s_{\psi_2}^{q_\theta}(x_t,t) - \nabla_{x_t} \log k(x_t|x_0)\|_2^2\right]. \label{eq:dsm_loss:q}
\end{align}
Thus, the gradient of DiKL w.r.t. the parameters $\theta$ of the student model can be estimated as follows, a method known as Variational Score Distillation (VSD)~\cite{pooledreamfusion,wang2024prolificdreamer,luo2024diff}:
\begin{align}
    \nabla_\theta \mathrm{DiKL}(q_\theta (x_0) || p_d(x_0)) 
    \approx \sum_{t=1}^T w(t) \int q_\theta(x_t) \big(s^{q_\theta}_{\psi_2}(x_t,t) - s^{p_d}_{\psi_1}(x_t,t)\big) \frac{\partial x_t}{\partial \theta} dx_t.  \label{eq:kl_gradient}
\end{align}
Unlike the teacher score function which remains fixed after pre-training, the noisy model score $\nabla_{x_t} \log q_\theta(x_t)$ dynamically changes as we update the student model's parameters $\theta$ during training. Therefore, the score network $s_{\psi_2}^{q_\theta}(x_t,t) \approx \nabla_{x_t} \log q_\theta(x_t)$ needs to be updated every time we update the student model, which results in an interleaved training procedure as detailed in Algorithm~\ref{alg:one_step_train}.

\textbf{Limitations of the VSD.}
Despite the initial success of VSD in diffusion distillation, this two-stage pipeline has inherent limitations.
In the first stage, the teacher score model $s_{\psi_1}^{p_d}(x_t,t)$ is imperfect due to limited network capacity and optimization.
In the second stage, the distillation process solely relies on the pre-trained teacher score for supervision.
Therefore, the gradient estimation in VSD (Eq.~\ref{eq:kl_gradient}) has biases from \emph{two sources}: (1)~the teacher's score estimation error, and (2)~the student's score estimation error.
These biases can accumulate and thus adversely affect the quality of the distilled student model.
In the next section, we propose a new method that directly estimates the score difference $\nabla_{x_t} \log q_\theta(x_t) - \nabla_{x_t} \log p_d(x_t)$ with a single lightweight network, which reduces the gradient estimation bias and improves generation quality.

\section{Training One-Step Diffusion Models with DiffRatio}\label{sec:method}
From Eq.~\ref{eq:kl:gradient}, we observe that the DiKL gradient depends on the difference between two score functions, which can be expressed as the gradient of a log density ratio:
\begin{align}
    \nabla_{x_t} \log q_\theta(x_t) - \nabla_{x_t} \log p_d(x_t)
    = \nabla_{x_t} \log \frac{q_\theta(x_t)}{p_d(x_t)}.
\end{align}
Therefore, instead of estimating the two score functions separately, we can directly learn the density ratio $\frac{q_\theta(x_t)}{p_d(x_t)}$ and then take the gradient of its logarithm to estimate the score difference. We thus refer to this approach as \textbf{\textsc{DiffRatio}}, which we introduce below.

\subsection{Diffusive Density Ratio Gradient Estimation}
We propose to use class ratio estimation~\cite{sugiyama2012density,qin1998inferences,gutmann2010noise,zhang2022out} to estimate the density ratio across all noise levels.
We first denote distributions \( q_{\theta}(x_t) \) and \( p_{d}(x_t) \) as two conditional distributions \( m(x_t|y=0) \) and \( m(x_t|y=1) \), respectively, where $y=0$ indicates samples from the one-step model $q_{\theta}(x_t)$ and $y=1$ indicates data samples from $p_{d}(x_t)$. With Bayes' rule, we can transform the density ratio estimation problem into a binary classification problem:
\begin{align}
    \frac{q_{\theta}(x_t)}{p_{d}(x_t)} &\equiv \frac{m(x_t|y=0)}{m(x_t|y=1)} = \frac{p(y=0|x_t) \cancel{m(x_t)}/\cancel{p(y=0)} }{ p(y=1|x_t) \cancel{m(x_t)} / \cancel{p(y=1)}} 
    = \frac{p(y=0|x_t)}{p(y=1|x_t)}, \label{eq:class:ratio}
\end{align}
where the mixture distribution is defined as $m(x)\equiv m(x_t|y=1) p(y=1) + m(x_t|y=0) p(y=0)$,
and the Bernoulli prior distribution \( p(y) \) is simply set as a uniform prior \( p(y=1) = p(y=0) = 0.5 \). In practice, this is achieved by drawing a batch of data from \( p_d(x_t) \) with the label \( y = 1 \) and drawing an equal number of samples from \( q_\theta(x_t) \) with the label \( y = 0 \). We then train a neural network classifier \( c_{\eta}(x_t, t) \), conditioned on the diffusion time \( t \), to estimate the probability that a given input \( x_t \) belongs to class \( y = 1 \). The optimal classifier approximates the posterior probability \( c_*(x_t,t)=p(y=1 | x_t) \). In this case, the log-density ratio can be written as
\begin{align}
    \nabla_{x_t} \log \frac{q_\theta(x_t)}{p_d(x_t)} 
    = \nabla_{x_t} \log \frac{1 - c_*(x_t, t)}{c_*(x_t, t)}
    = \nabla_{x_t} \operatorname{logit}(1 - c_*(x_t, t)) \label{eq:class_ratio_est_dikl}.
\end{align}
We can then obtain a new DiKL gradient estimator by plugging in the density ratio estimator to Eq.~\ref{eq:kl:gradient}:
\begin{align}
    \nabla_\theta \mathrm{DiKL}(q_\theta (x_0) || p_d(x_0)) 
    \approx \sum_{t=1}^T w(t) \int q_\theta(x_t)  \nabla_{x_t} \text{logit}(1 - c_\eta(x_t, t)) \frac{\partial x_t}{\partial \theta} dx_t.\label{eq:dikl:cle}
\end{align}
Importantly, our method directly estimates the score difference $\nabla_{x_t} \log q_\theta(x_t) - \nabla_{x_t} \log p_d(x_t)$ using a single density ratio estimator $c_\eta(x_t,t)$, which alleviates the two sources of bias in VSD (Eq.~\ref{eq:kl_gradient}):
(1)~Instead of separately estimating the student score $s_{\psi_2}^{q_\theta}(x_t,t)$ and the teacher score $s_{\psi_1}^{p_d}(x_t,t)$, the density ratio estimator directly estimates the score difference in an integrated manner, avoiding the accumulation of errors from two independently trained networks.
(2)~Without relying on a teacher score for supervision, we avoid bias propagation from the teacher to the student, leading to a \emph{consistent} estimator---given sufficient capacity and training, the estimated score difference converges to the true score difference. In contrast, \emph{VSD cannot guarantee such consistency}, as the bias from teacher supervision cannot be removed regardless of how well the student score network is trained. In the next section, we give an empirical comparison of the gradient estimation bias between VSD and DiffRatio methods.

\begin{figure}[t]
    \centering

    \begin{minipage}[t]{0.58\linewidth}
        \vspace{0pt}
        \centering
        \includegraphics[width=\linewidth]{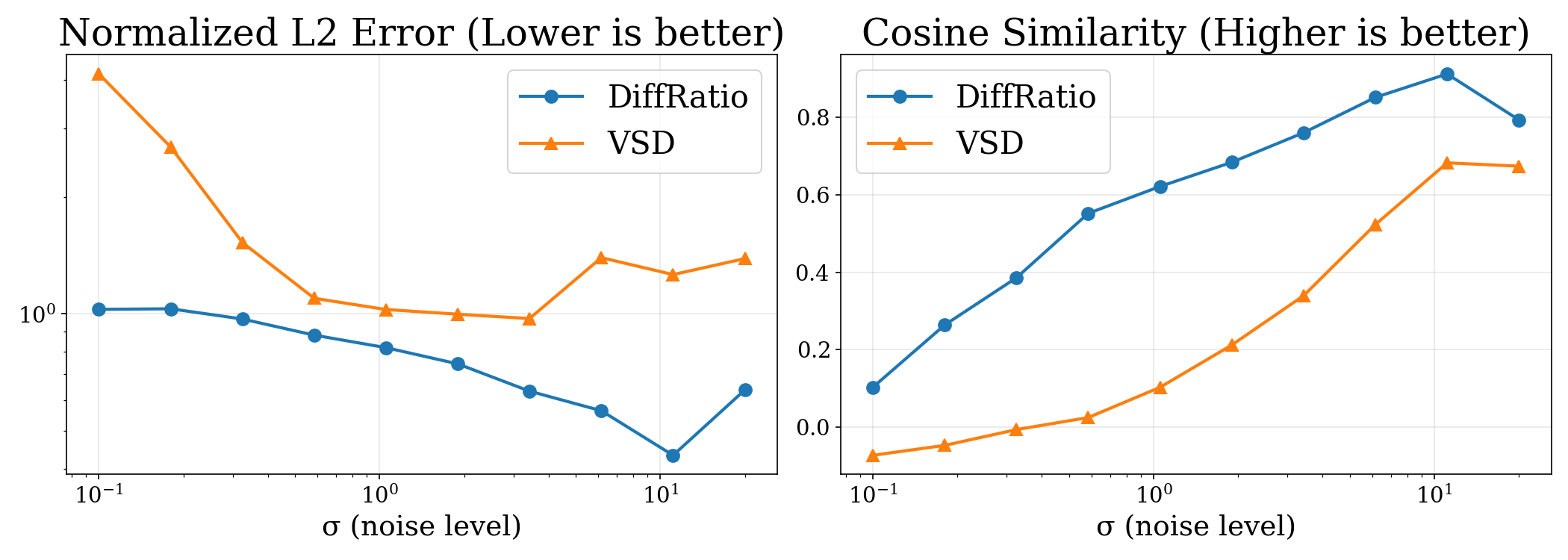}
        \caption{Score difference estimation accuracy on a 2D mixture of Gaussians problem.}
        \label{fig:toy_comparison}
    \end{minipage}
    \hfill
    \begin{minipage}[t]{0.35\linewidth}
        \vspace{0pt}
        \centering
        \small
        \captionof{table}{Evaluation of trained generators on the 2D toy problem. Log-MMD ($\downarrow$): lower is better. Log-density ($\uparrow$): higher is better.}
        \label{tab:toy_eval}
        \begin{tabular}{@{}l|cc@{}}
            \toprule
            \textsc{Method} & Log-MMD & Log-density \\
            \midrule
            True & / & $-6.65$ \\
            VSD & $-5.25$ & $-10.88$ \\
            DiffRatio & $\mathbf{-6.08}$ & $\mathbf{-6.79}$  \\
            \bottomrule
        \end{tabular}
    \end{minipage}
\end{figure}

\subsection{Empirical Analysis of the Gradient Bias}\label{sec:toy}

To empirically validate that our density-ratio-based gradient estimator reduces bias compared to VSD, we design a controlled 2D experiment where ground-truth score differences can be computed.
The data distribution $p_d(x_0)$ is a 2D mixture of Gaussians with 40 components, which allows closed-form computation of the true data score $\nabla_{x_t} \log p_d(x_t)$ at any noise level.
The one-step model $q_\theta(x_0)$ is an implicit generative model that has been pre-trained to approximate $p_d$. Since $q_\theta(x_t)$ lacks a closed-form density, we estimate its score using kernel density estimation (KDE) with 10,000 samples, which provides highly accurate score estimates in this low-dimensional setting. 
This gives us the ground-truth score difference: $\Delta s_t^*\equiv\Delta s^*(x_t, t) = \nabla_{x_t} \log q_\theta(x_t) - \nabla_{x_t} \log p_d(x_t)$.
We compare the following two methods on this problem: (1) DiffRatio (ours) trains a single classifier $c_\eta(x_t, t)$ and estimates the score difference as $\widehat{\Delta s_t} = \nabla_{x_t} \textrm{logit}(1 - c_\eta(x_t, t))$; (2) VSD trains two separate score networks $s_{\psi_1}^{p_d}$ and $s_{\psi_2}^{q_\theta}$ using DSM and computes their difference $\widehat{\Delta s_t} = s_{\psi_2}^{q_\theta}(x_t, t) - s_{\psi_1}^{p_d}(x_t, t)$.

We evaluate the performance using two metrics: (1)~the relative L2 error $\|\widehat{\Delta s_t} - \Delta s_t^*\|_2 / \|\Delta s_t^*\|_2$, and (2)~cosine similarity, where $\widehat{\Delta s_t}$ denotes the estimated score difference from each method. We train both methods with 5,000 steps; other details can be found in  Appendix~\ref{appendix:toy_setup}.
As shown in Figure~\ref{fig:toy_comparison}, our density-ratio-based method consistently achieves lower L2 error and higher cosine similarity than VSD across all noise levels, indicating more accurate gradient estimation. 
This is because VSD accumulates errors from both the student score network $s_{\psi_2}^{q_\theta}$ and the teacher score network $s_{\psi_1}^{p_d}$. 
In contrast, our method directly estimates the score \emph{difference} using a single density ratio estimator $c_\eta$, avoiding error accumulation from separately trained networks. 
This empirically validates that density ratio estimation provides more accurate gradient signals for training one-step models.
To further illustrate the effect of gradient bias to generation quality, we compare the two gradient estimation methods by training one-step models under the same divergence (DiKL) on this problem.
We report log-MMD (maximum mean discrepancy) and log-density under the true data distribution in Table~\ref{tab:toy_eval}.  Our method achieves significantly better log-density than VSD, indicating that the generated samples are closer to the true data distribution. The improved gradient estimation translates directly to better generation quality. 

\subsection{Training Criterion Extensions}
In addition to the DiKL, we can use the learned classifier function \( c_\eta \) to obtain a family of gradient estimators for alternative training objectives. For instance, replacing the logit function with the logarithm function yields an objective that minimizes the probability of generated samples being classified as fake. This formulation aligns with GAN~\cite{goodfellow2014generative, nowozin2016f} across different diffusion time steps, which approximately minimizes the \emph{Diffusive Jensen-Shannon (DiJS)} divergence (a detailed derivation of DiJS and its gradient estimator can be found in Appendix~\ref{appendix:dijs}): 
\begin{align}
    \nabla_\theta \mathrm{DiJS}(q_\theta (x_0) || p_d(x_0))
    \approx \sum_{t=1}^T w(t) \int q_\theta(x_t)  \nabla_{x_t} \log (1 - c_\eta(x_t, t)) \frac{\partial x_t}{\partial \theta} dx_t. \label{eq:diJS}
\end{align}
Interestingly, the DiJS objective mirrors the minimax GAN formulation, which minimizes the probability of the generated images being fake. Alternatively, one can also maximize the probability of them being real. This approach is referred to as \emph{Diffusive Realism Maximization (DiRM)}, which has the following gradient estimator: 
\begin{align}
    \nabla_\theta \mathrm{DiRM}(\theta) 
    \approx -\sum_{t=1}^T w(t) \int q_\theta(x_t)  \nabla_{x_t} \log c_\eta(x_t, t) \frac{\partial x_t}{\partial \theta} dx_t. \label{eq:diRM}
\end{align}
Notably, the DiRM objective, which maximizes the likelihood of the generated images being real, mirrors the \emph{non-saturating GAN} formulation~\cite{goodfellow2014generative}, which is known to provide more stable gradients for the generator compared to the original minimax objective. 

In principle, once the density ratio is available, any ratio-based divergence (e.g., $f$-divergence \cite{xu2025one}) can be used for training. This flexibility is not available when one only has access to the teacher and student scores (as in VSD), making DiffRatio applicable to a broader family of criteria. Moreover, we provide a convergence discussion in Appendix~\ref{appendix:convergence}, where we justify the consistency of our DiffRatio gradient estimators under \emph{weaker assumptions} than those required for standard ratio estimation~\cite{hyvarinen2005estimation}.

\begin{algorithm}[t]
    \caption{Training one-step diffusion models via DiffRatio (ours)}
    \label{alg:one_step_train:score:free}
    \begin{algorithmic}[1]
        \Require Training samples $\mathcal{D} \sim p_d(x_0)$
        \Statex \textcolor{gray}{\textit{// Stage 1: Pre-train a score model for initialization}}
        \State Pre-train a score network $s_{\theta}^{p_d}(x_t,t)$ using DSM (Eq.~\ref{eq:dsm_loss:pd}) until convergence
        \Statex \textcolor{gray}{\textit{// Stage 2: Train the one-step model without teacher supervision}}
        \State Create a one-step generator $g_{\theta}(\cdot)\equiv s_{\theta}^{p_d}(\cdot,t=t_{\text{init}})$ by fixing the diffusion time argument $t=t_{\text{init}}$ in the pretrained score network
       \For{each training iteration}
            \State Sample time $t'$ for diffusive density ratio estimation
            \State Train a classifier $c_\eta(x_{t'},t')$ on samples from $q_\theta(x_{t'}),p_d(x_{t'})$ to estimate their density ratio
            \State Update the one-step generator $g_\theta$ with $c_\eta(x_{t'},t')$ using Eq.~\ref{eq:dikl:cle} or Eq.~\ref{eq:diJS} or Eq.~\ref{eq:diRM}
        \EndFor
    \end{algorithmic}
\end{algorithm}

\subsection{Simplified Training Procedure}

The overall training procedure of our proposed DiffRatio framework is summarized in Algorithm~\ref{alg:one_step_train:score:free}. In our method, the generator is first initialized with a pre-trained score network and then further trained as a one-step model with the diffusion time argument in the score network fixed to $t{=}t_{\text{init}}$. This design leads to a more simplified training pipeline compared to VSD. Specifically, VSD requires maintaining and evaluating three networks during training: (1)~the pre-trained teacher score network $s_{\psi_1}^{p_d}$, (2)~the student score network $s_{\psi_2}^{q_\theta}$, and (3)~the generator $g_\theta$. In contrast, our method only requires two: (1)~the generator $g_\theta$, and (2)~the density ratio estimator $c_\eta$ (which is approximately half the size of a full score network). This significantly reduces the size of the auxiliary networks during training.

For low-dimensional problems such as the 2D mixture of Gaussians in Section~\ref{sec:toy}, the generator can be trained from random initialization. However, for high-dimensional image generation, we find that initializing the generator with a pre-trained score network is necessary to prevent mode collapse, which is consistent with findings in the score-based distillation literature~\cite{luo2024diff,zhou2024adversarial,zhou2024score}. To further understand the necessity of this pre-training, we provide a detailed analysis in Section~\ref{sec:init}, which studies the role of weight initialization from two perspectives: (1)~the functional mapping perspective, and (2)~the feature space perspective. Our analysis shows that multi-level features learned during diffusion pre-training are essential for preventing mode collapse in high-dimensional settings.

\subsection{Inference-Time Parallel Scaling with Learned Density Ratio Network}\label{sec:scaling}
Unlike existing distillation approaches \cite{luo2024diff,zhou2024adversarial,zhou2024score} where the learned student score $s_{\psi_1}^{q_\theta}(x_t,t)$ network is rarely useful after training, the learned density ratio network in our approach can be leveraged at inference time as a judge to further refine generation quality.
Specifically, to make the samples from the trained model $q_\theta(x)$ closer to samples from $p_d(x)$, we noticed that 
\begin{align}
p_d(x_0)=q_\theta(x_0)\frac{p_d(x_0)}{q_\theta(x_0)}= q_\theta(x_0) \frac{c_*(x_0)}{1-c_*(x_0)},
\end{align}
which allows us to perform \emph{importance resampling} to correct the samples with our consistent ratio estimator. Similar constructions have been explored in the GAN–EBM literature~\cite{che2020your,arbel2020generalized}.

In practice, we first generate $M$ candidates from our one-step model: $x_0^{(1:M)} \sim q_\theta(x_0)$, then we select one sample $x_0^{(i)}$ from $\{x_0^{(1:M)}\}$ according to the categorical distribution
\begin{equation}
i \sim \mathrm{Cat}\!\left(
\frac{w(x_0^{(1)})}{\sum_{j=1}^M w(x_0^{(j)})}, \cdots, \frac{w(x_0^{(M)})}{\sum_{j=1}^M w(x_0^{(j)})}\right),
\end{equation}
where the density ratio $w$ is directly approximated by the learned ratio  at $t_{\text{min}}$ to avoid the indefiniteness of the density ratio for manifold distributions~\cite{song2019generative}:
\begin{equation}
    w(x_0^{(i)})\equiv\frac{p_d(x_0^{(i)})}{q_\theta(x_0^{(i)})}\approx\frac{c_\eta(x_{t_\text{min}}^{(i)},t_\text{min})}{1-c_\eta(x_{t_\text{min}}^{(i)},t_\text{min})}. \label{eq:density_ratio}
\end{equation}
 This procedure enables effective inference-time scaling: increasing $M$ monotonically reduces the sampling bias of the importance-resampled estimator of the target distribution. Crucially, all $M$ samples and their corresponding ratio evaluations can be computed fully \emph{in parallel}, making this inference-time scaling step highly efficient in practice.

 \section{Related Work}
\label{sec:gan}
Density ratio estimation lies at the core of many GAN variants~\cite{goodfellow2014generative,nowozin2016f}. In classic GANs, the discriminator implicitly estimates the density ratio between the data and model distributions. However, in high-dimensional image modeling, both distributions typically concentrate on low-dimensional manifolds and are not absolutely continuous, making the density ratio—and consequently the Jensen--Shannon divergence—ill-defined. This issue is widely recognized as a major source of GAN training instability~\cite{arjovsky2017towards,arjovsky2017wasserstein,mescheder2018training,roth2017stabilizing}.
To address this, Wasserstein GANs replace the JS divergence with the Wasserstein-1 distance, which remains well-defined even when the distributions are disjoint~\cite{arjovsky2017wasserstein}. In practice, enforcing the required 1-Lipschitz constraint is challenging and relies on heuristic regularization methods such as weight clipping~\cite{arjovsky2017wasserstein}, gradient penalties~\cite{gulrajani2017improved}, or spectral normalization~\cite{miyato2018spectral}. As a result, the optimized objective often deviates from the ideal divergence, and empirical stability is largely driven by regularization rather than strict divergence minimization~\cite{mescheder2018training,fedus2018many}.

Adding Gaussian noise to real and fake samples has been proposed as a way to ensure distributions are fully supported, making the density ratio well-defined~\cite{sonderby2016amortised,roth2017stabilizing,nowozin2016f,zhang2020spread}. This model-agnostic approach requires no architectural changes but hinges on choosing an effective noise level—something hard to fix throughout training.  
Diffusion GAN~\cite{wang2022diffusion} addresses the challenge of selecting a fixed noise level by introducing a diffusion-inspired noise schedule that gradually increases noise in tandem with the model’s learning capacity. While this represents the most closely related work to ours, our approach differs in some important aspects.

First, DiffRatio is agnostic to the generator's network architecture.
Diffusion GAN stabilizes training using a StyleGAN-based generator~\cite{karras2024analyzing}, which is implicitly trained progressively from low to high resolutions while keeping the network topology fixed. 
In contrast, DiffRatio does not rely on a specialized generator architecture or progressive resolution training. Instead, we adopt a generic U-Net architecture, resulting in a simpler and more broadly applicable framework.
Second, DiffRatio does not rely on additional regularization techniques or GAN-specific training tricks such as gradient penalties or spectral normalization. Therefore, we provide a clean framework for training one-step diffusion models yet still achieves stable convergence and demonstrates training robustness.

Therefore, our method leads to a conceptually cleaner training objective that is more amenable to theoretical analysis and empirical diagnosis, as it avoids the complexities and instabilities associated with specific network architectures or additional regularization tricks.

\section{Experiments}\label{sec:experiments}
\vspace{-1ex}

\begin{table*}[t]
    \centering

    \begin{minipage}[t]{0.42\textwidth}
        \vspace{0pt}
        \scriptsize
        \centering
        \captionof{table}{Unconditional one-step generation on CIFAR-10.}
        \label{tab:cifar-10-generation-result}
        \setlength{\tabcolsep}{4pt}
        \resizebox{\linewidth}{!}{%
        \begin{tabular}{@{}lccc@{}}
            \toprule
            \textsc{METHOD} & \textsc{NFE} ($\downarrow$) & \textsc{FID} ($\downarrow$) & \textsc{IS} ($\uparrow$) \\
            \midrule
            \multicolumn{4}{@{}l}{\textbf{Teacher model}} \\
            \midrule
            EDM \cite{karras2022elucidating} & 35 & 2.04 & 9.84 \\
            EDM \cite{karras2022elucidating} & 1 & 8.70 & 8.49 \\
            \midrule
            \multicolumn{4}{@{}l}{\textbf{Training from scratch without teacher}} \\
            \midrule
            CT \cite{song2023consistency} & 1 & 8.70 & 8.49 \\
            iCT \cite{songimproved} & 1 & 2.83 & 9.54 \\
            iCT-deep \cite{songimproved} & 1 & \textbf{2.51} & \textbf{9.76} \\
            BCM \cite{li2024bidirectional} & 1 & 3.10 & 9.45 \\
            BCM-deep \cite{li2024bidirectional} & 1 & 2.64 & 9.67 \\
            sCT \cite{lu2024simplifying} & 1 & 2.85 & - \\
            Diffusion-GAN \cite{wang2022diffusion} & 1 & 3.19 & - \\
            IMM \cite{zhou2025inductive} & 1 & 3.20 & - \\
            MeanFlow \cite{geng2025mean} & 1 & 2.92 & - \\
            \midrule
            \multicolumn{4}{@{}l}{\textbf{Distillation with teacher supervision}} \\
            \midrule
            Progressive Distillation \cite{salimans2022progressive} & 1 & 8.34 & 8.69 \\
            DSNO \cite{zheng2023fast} & 1 & 3.78 & - \\
            TRACT \cite{berthelot2023tract} & 1 & 3.78 & - \\
            CD \cite{song2023consistency} & 1 & 3.55 & 9.48 \\
            CTM (w/ GAN) \cite{kim2023consistency} & 1 & 1.98 & - \\
            CTM (w/o GAN) \cite{kim2023consistency} & 1 & ${>}$5.0 & - \\
            sCD \cite{lu2024simplifying} & 1 & 3.66 & - \\
            Diff-Instruct \cite{luo2024diff} & 1 & 4.53 & - \\
            ECT \cite{geng2025consistency} & 1 & 3.60 & - \\
            SiD \cite{zhou2024score} & 1 & 2.03 & 10.02 \\
            SiDA \cite{zhou2024adversarial} & 1 & \textbf{1.52} & \textbf{10.32} \\
            \midrule
            \multicolumn{4}{@{}l}{\textbf{Training without teacher supervision (ours)}} \\
            \midrule
            DiffRatio-DiRM & 1 & 4.87 & 9.85 \\
            DiffRatio-DiKL & 1 & 3.81 & 9.90 \\
            DiffRatio-DiJS & 1 & \textbf{2.39} & \textbf{9.93} \\
            \bottomrule
        \end{tabular}
        }
    \end{minipage}
    \hfill
    \begin{minipage}[t]{0.415\textwidth}
        \vspace{0pt}
        \scriptsize
        \centering
        \captionof{table}{Conditional one-step generation on ImageNet $64{\times}64$.}
        \label{tab:imagenet64-generation-result}
        \setlength{\tabcolsep}{5pt}
        \resizebox{\linewidth}{!}{%
        \begin{tabular}{@{}lcc@{}}
            \toprule
            \textsc{METHOD} & \textsc{NFE} ($\downarrow$) & \textsc{FID} ($\downarrow$) \\
            \midrule
            \multicolumn{3}{@{}l}{\textbf{Teacher model}} \\
            \midrule
            EDM \cite{karras2022elucidating} & 79 & 2.44 \\
            \midrule
            \multicolumn{3}{@{}l}{\textbf{Training from scratch without teacher}} \\
            \midrule
            EDM2-L/XL \cite{karras2024analyzing} & 1 & 13.0 \\
            CT \cite{song2023consistency} & 1 & 13.0 \\
            iCT \cite{songimproved} & 1 & 4.02 \\
            iCT-deep \cite{songimproved} & 1 & 3.25 \\
            BCM \cite{li2024bidirectional} & 1 & 4.18 \\
            BCM-deep \cite{li2024bidirectional} & 1 & 3.14 \\
            sCT \cite{lu2024simplifying} & 1 & \textbf{2.04} \\
            \midrule
            \multicolumn{3}{@{}l}{\textbf{Distillation with teacher supervision}} \\
            \midrule
            Progressive Distillation \cite{salimans2022progressive} & 1 & 7.88 \\
            DSNO \cite{zheng2023fast} & 1 & 7.83 \\
            TRACT \cite{berthelot2023tract} & 1 & 7.43 \\
            CD \cite{song2023consistency} & 1 & 6.20 \\
            CTM (w/ GAN) \cite{kim2023consistency} & 1 & 1.92 \\
            sCD \cite{lu2024simplifying} & 1 & 2.44 \\
            Diff-Instruct \cite{luo2024diff} & 1 & 5.57 \\
            EM Distillation \cite{xie2024distillation} & 1 & 2.20 \\
            ECT \cite{geng2025consistency} & 1 & 2.49 \\
            SiD \cite{zhou2024score} & 1 & 2.02 \\
            SiDA \cite{zhou2024adversarial} & 1 & \textbf{1.35} \\
            DMD2 (w/ GAN) \cite{yin2024improved} & 1 & 1.51 \\
            \midrule
            \multicolumn{3}{@{}l}{\textbf{Training without teacher supervision (ours)}} \\
            \midrule
            DiffRatio-DiJS & 1 & \textbf{1.54} \\
            \bottomrule
        \end{tabular}
        }
    \end{minipage}
    \vspace{-2ex}
\end{table*}

We evaluate the performance of our proposed method by training one-step image diffusion models on two standard benchmarks: (1) unconditional generation on the CIFAR-10 ($32{\times}32$) dataset~\cite{krizhevsky2009learning}, (2) class-label-conditioned generation on the ImageNet ($64{\times}64$ and $512{\times}512$) dataset~\cite{deng2009imagenet}. 
Our one-step diffusion model adopts the same neural network architecture as the pre-trained score model. Specifically, we employ the EDM U-Net model architecture \cite{karras2022elucidating} for CIFAR-10 and ImageNet $64{\times}64$, and use the EDM2-M U-Net model architecture \cite{karras2024analyzing} for ImageNet $512{\times}512$. We consider three training criteria in our framework (DiKL/DiJS/DiRM) on CIFAR-10, and use the best-performing criterion (DiJS) for higher-dimensional ImageNet experiments ($64{\times}64$ and $512{\times}512$). We use the variance-exploding (VE) noise schedule to define the diffusive divergences.
Our density ratio estimator \( c_\eta(x_t, t) \) is implemented using the encoder portion of the teacher's U-Net architecture with a sigmoid function attached to the end to produce a scalar output between $0$ and $1$. This network is approximately half the size of the full U-Net used for score estimation, leading to reduced size of auxiliary networks. The one-step generator \( g_\theta(z) \) is initialized using the weights from the pre-trained score model with the diffusion time argument $t_{\text{init}}{=}2.5$ fixed throughout training and sampling. 
We follow standard hyperparameter settings for training generative models on CIFAR-10 and ImageNet as detailed in \cite{karras2022elucidating,karras2024analyzing}. 
We use a batch size of 64 for CIFAR-10 and 4,096 for ImageNet.
The weighting function \( w(t) {=} \sigma_t^2 \) is used for for both datasets. 
We adopt a single-step update strategy for the density ratio estimator throughout our experiments, consistent with previous works~\cite{luo2024diff,zhou2024score}.
See Appendix~\ref{appendix:exp_setup} for more details.

\begin{figure*}[t]
    \centering

    \begin{minipage}[t]{0.36\textwidth}
        \vspace{0pt}
    \centering

    \includegraphics[width=0.48\linewidth]{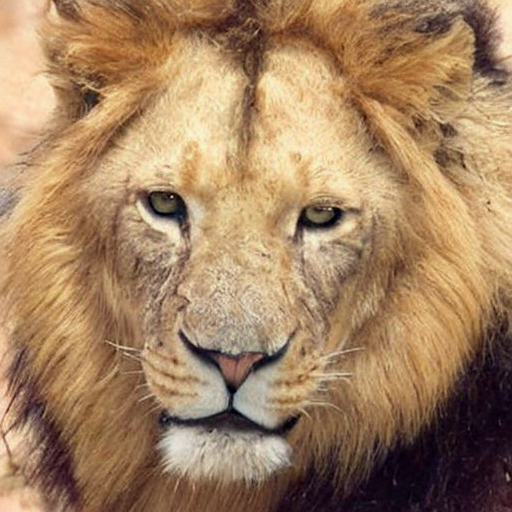}\hfill
    \includegraphics[width=0.48\linewidth]{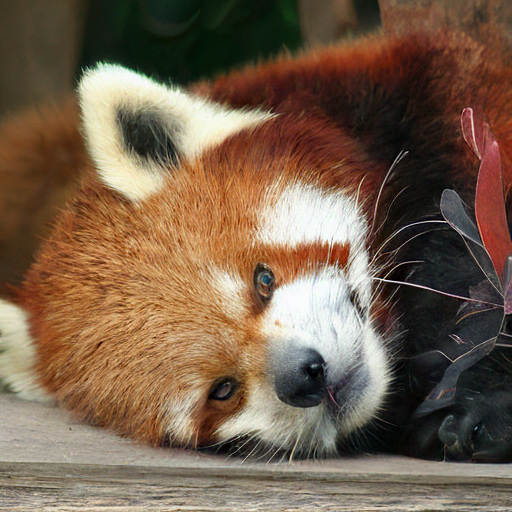}\par\vspace{2pt}

    \includegraphics[width=0.48\linewidth]{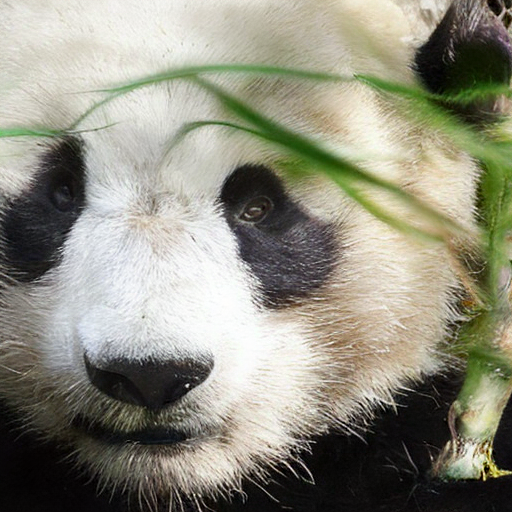}\hfill
    \includegraphics[width=0.48\linewidth]{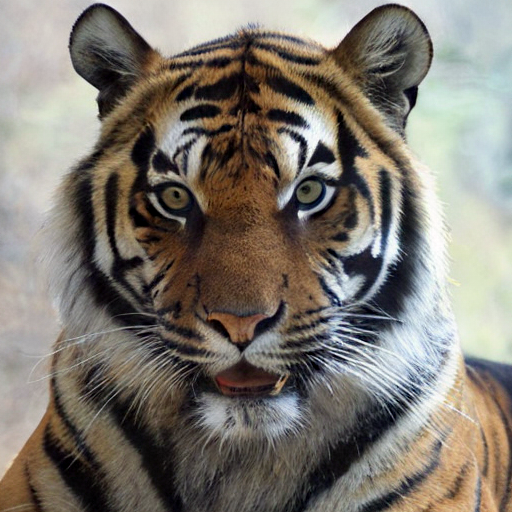}\par\vspace{2pt}

    \includegraphics[width=0.48\linewidth]{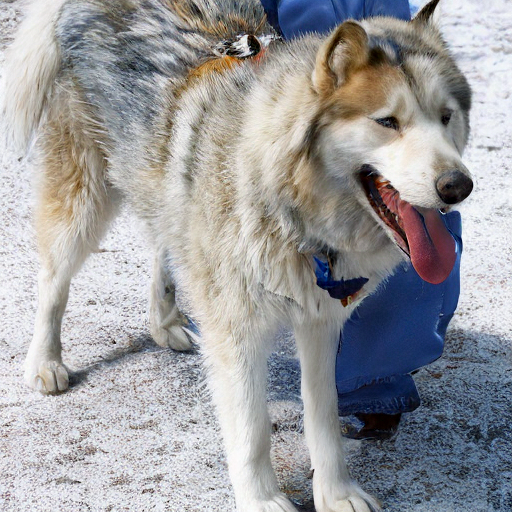}\hfill
    \includegraphics[width=0.48\linewidth]{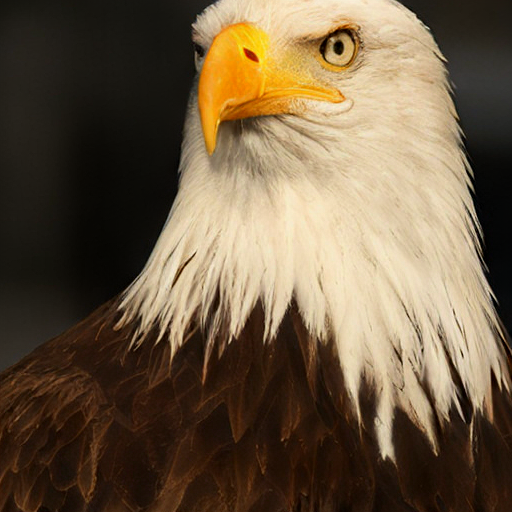}\par\vspace{2pt}

        \captionof{figure}{Images generated by DiffRatio-DiJS-M on ImageNet $512{\times}512$ (FID=1.41) with 1 NFE.}
        \label{fig:imagenet512_samples}
    \end{minipage}
    \hfill
    \begin{minipage}[t]{0.6\textwidth}
        \vspace{0pt}
        \scriptsize
        \centering
        \captionof{table}{Conditional one-step generation on ImageNet $512{\times}512$. $\dagger$ indicates that NFE is executed \emph{in parallel} rather than sequentially.}
        \label{tab:imagenet512-generation-result}
        \setlength{\tabcolsep}{5pt}

        \resizebox{\linewidth}{!}{%
        \begin{tabular}{@{}lccc@{}}
            \toprule
            \textsc{METHOD} & CFG & \textsc{NFE} ($\downarrow$) & \textsc{FID} ($\downarrow$) \\
            \midrule
            \multicolumn{4}{@{}l}{\textbf{Teacher model}} \\
            \midrule
            EDM2-M \cite{karras2024analyzing} & N & 63 & 2.25 \\
            EDM2-M \cite{karras2024analyzing} & Y & 126 & 2.01 \\
            \midrule
            \multicolumn{4}{@{}l}{\textbf{Training from scratch without teacher}} \\
            \midrule
            sCT-M \cite{lu2024simplifying} & Y & 1 & 5.84 \\
            sCT-M \cite{lu2024simplifying} & Y & 2 & 5.53 \\
            sCD-M \cite{lu2024simplifying} & Y & 1 & \textbf{2.75} \\
            sCD-M \cite{lu2024simplifying} & Y & 2 & \textbf{2.26} \\
            \midrule
            \multicolumn{4}{@{}l}{\textbf{Distillation with teacher supervision}} \\
            \midrule
            SiD-M \cite{zhou2024score} & N & 1 & 2.06 \\
            SiDA-M \cite{zhou2024adversarial} & N & 1 & \textbf{1.55} \\
            \midrule
            \multicolumn{4}{@{}l}{\textbf{Training without teacher supervision (ours)}} \\
            \midrule
            DiffRatio-DiJS-M & N & 1 & \textbf{1.41} \\
            DiffRatio-DiJS-M (w/ inference-time scaling) & N & 10$^\dagger$ & 1.35 \\
            DiffRatio-DiJS-M (w/ inference-time scaling) & N & 20$^\dagger$ & 1.34 \\
            DiffRatio-DiJS-M (w/ inference-time scaling) & N & 50$^\dagger$ & 1.34 \\
            DiffRatio-DiJS-M (w/ inference-time scaling) & N & 100$^\dagger$ & \textbf{1.32} \\
            \bottomrule
        \end{tabular}
        }
    \end{minipage}
    \vspace{-2.5ex}
\end{figure*}

In Tables~\ref{tab:cifar-10-generation-result}, \ref{tab:imagenet64-generation-result} and \ref{tab:imagenet512-generation-result}, we compare our method to previous methods for training one-step generative models. To highlight methodological differences, we categorize these approaches into three groups:  
(1) training from scratch,  
(2) distillation with teacher supervision (e.g., using teacher score, denoiser or ODE trajectory as supervision in the distillation loss), and  
(3) training without teacher supervision. 
We find that our proposed method, DiffRatio, achieves competitive one-step generation performance despite not using any teacher supervision, comparable to most state-of-the-art distillation methods that rely on full teacher supervision. The only baseline method with similar performance to ours is SiDA~\cite{zhou2024adversarial}, which depends on training data, teacher score supervision, teacher weight initialization, and student score estimation. In contrast, our method requires only training data, teacher weight initialization, and density ratio estimation. Notably, this \emph{eliminates} the need for teacher supervision in the training process. Also, density ratio estimation is both simpler and \emph{more lightweight} than student score estimation, as the density ratio network is approximately half the size of a full score network. Moreover, unlike VSD, our approach does not require maintaining or evaluating the teacher score network during the training process. This results in a more streamlined training framework, since our approach does not require maintaining and evaluating the teacher and student score networks in the training procedure. Images generated by DiffRatio-DiJS are visualized in Figures~\ref{fig:imagenet512_samples}. More generated images can be found in Appendix \ref{appendix:additional_images} (Figures~ \ref{fig:dijs_cifar10_full}, \ref{fig:dijs_imagenet64_full} and \ref{fig:dijs_imagenet512_full}).

\textbf{Inference-Time Parallel Scaling.}
We further explore the inference-time refinement of our approach using the learned ratio, as we discussed in \Cref{sec:scaling}.
We first investigate the following question: \emph{after training, is there still a discrepancy between the distributions of the real and generated data?}
\begin{wrapfigure}{r}{0.35\linewidth}
    \centering
    \vspace{-1ex}
    \includegraphics[width=\linewidth]{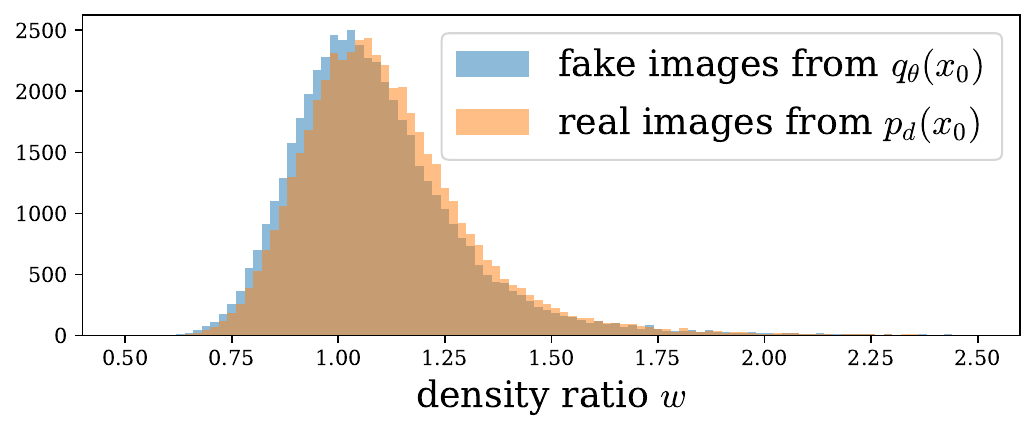}
    \caption{Histograms of the learned density ratio $w$.}
    \label{fig:density_ratio_hist}
    \vspace{-3ex}
\end{wrapfigure}
We evaluate the learned density ratio on both real and generated images by Eq.~\ref{eq:density_ratio}
and visualize the resulting distributions via histograms in Figure~\ref{fig:density_ratio_hist}.
We can see that the two distributions largely overlap, indicating that our generator achieves a relatively good performance.
However, real images still exhibit a slightly larger ratio on average than generated ones, suggesting a small but consistent  mismatch between $q_\theta(x_0)$ and $p_d(x_0)$, captured by the learned ratio.
We therefore investigate inference-time scaling via importance resampling using the learned ratio on ImageNet $512{\times}512$.
In \Cref{tab:imagenet512-generation-result}, we report results with candidate set sizes $M \in \{10, 20, 50, 100\}$.
The FID decreases monotonically as we allocate more inference-time compute, resulting in a significant improvement from $1.41$ to $1.32$.

\section{Analysis of the Role of Weight Initialization in One-Step Diffusion Models}\label{sec:init}
\begin{figure*}[t]
    \centering

    \begin{minipage}[t]{0.65\textwidth}
        \vspace{0pt}
        \centering

        \begin{subfigure}[t]{0.32\linewidth}
            \centering
            \includegraphics[width=\linewidth]{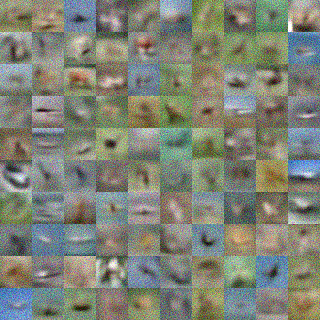}
            \caption{Single-level Init.}
            \label{fig:img1}
        \end{subfigure}
        \hfill
        \begin{subfigure}[t]{0.32\linewidth}
            \centering
            \includegraphics[width=\linewidth]{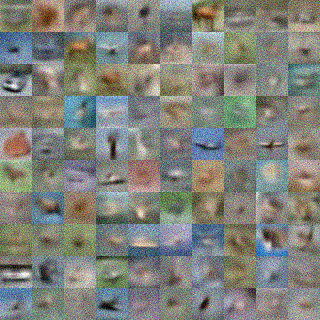}
            \caption{Multi-level Init.}
            \label{fig:img2}
        \end{subfigure}
        \hfill
        \begin{subfigure}[t]{0.32\linewidth}
            \centering
            \includegraphics[width=\linewidth]{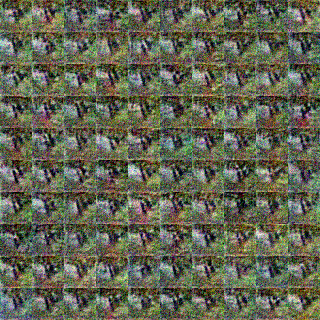}
            \caption{Collapsed Samples}
            \label{fig:img4}
        \end{subfigure}

        \captionof{figure}{
            Visualization of samples from models trained with different initializations
            and collapsed samples on CIFAR-10.
        }
        \label{fig:init_comparison}
    \end{minipage}
    \hfill
    \begin{minipage}[t]{0.33\textwidth}
        \vspace{0pt}
        \centering
        \tiny

        \captionof{table}{
            One-step models trained by DiJS with
            different initializations on various CIFAR subsets.
        }
        \label{tab:init_results}

        \setlength{\tabcolsep}{3pt}
        \renewcommand{\arraystretch}{1.12}

        \begin{tabularx}{\linewidth}{
            @{}
            >{\raggedright\arraybackslash}p{0.29\linewidth}
            >{\raggedright\arraybackslash}X
            >{\centering\arraybackslash}p{0.16\linewidth}
            @{}
        }
            \toprule
            Initialization & Initialization dataset & FID \\
            \midrule
            No init. & -- & collapsed \\
            \midrule
            Single-level DSM init. & full CIFAR-10 & collapsed \\
            \midrule
            \multirow{4}{=}{Multi-level DSM init.}
                & 10 classes in CIFAR-100 & collapsed \\
                & 50 classes in CIFAR-100 & 6.20 \\
                & 90 classes in CIFAR-100 & 6.01 \\
                & full CIFAR-10 & 2.39 \\
            \bottomrule
        \end{tabularx}
    \end{minipage}

\end{figure*}

Our one-step diffusion model was initialized with the pre-trained score model's weights.
We observed that training from random initialization led to mode collapse; see Figure~\ref{fig:img4} for an example of mode collapse. One possible explanation is that mode collapse arises from the training objectives (i.e., reverse KL or JS divergence), a phenomenon also observed in GAN literature~\cite{goodfellow2014generative}. To understand why initializing the one-step model with the pre-trained score model’s weights prevents mode collapse in the training process, we investigate the following two hypotheses.

\textbf{Function Space Hypothesis.} \emph{Teacher weight initialization provides a more structured latent-to-output functional mapping, i.e., different locations in the latent space are initially mapped to distinct images, preventing mode collapse.} 

This hypothesis originally arose from visualizing initialized samples as shown in Figure~\ref{fig:img1}, showing that initialization already induces diverse mappings, with the one-step model training stage primarily refining these initializations into sharper images. Somewhat surprisingly, however, we find that functional initialization alone is insufficient to prevent mode collapse. To show this, instead of training the teacher model across different diffusion time steps \( t \) and selecting a single time step \( t_{\text{init}} \) for initialization, we only pre-train the teacher model at the selected time step \( t_{\text{init}} \) and use its weight to initialize the one-step model. This setup ensures identical latent-to-output mappings for the one-step model at initialization as shown in Figure~\ref{fig:img2}. However, with this initialization, the one-step model still exhibits mode collapse early in the one-step model training stage, which suggests that the functional mapping perspective alone does not fully explain the mode-collapse issue.

\textbf{Feature Space Hypothesis.} \emph{Teacher weight initialization provides a rich set of multi-level features learned when pre-training the teacher diffusion model, which help prevent mode collapse.} 

To verify this hypothesis and isolate the role of learned features from functional mapping effects, we pre-train the teacher model on CIFAR-100 while excluding all classes that overlap with CIFAR-10. This ensures that images from the target classes that the one-step model aims to generate are absent during pre-training, allowing us to focus solely on the contribution of the learned features. We train the teacher model using increasingly larger subsets of CIFAR-100 with (10, 50, 90) classes, creating a setting with increasing feature diversity. Table~\ref{tab:init_results} shows the performance of our one-step model on CIFAR-10 initialized with the weights of teacher models trained on varying numbers of CIFAR-100 classes. We find that when the teacher model is trained on only 10 classes, mode collapse still occurs. However, as the number of training classes increases, the one-step model no longer collapses, indicating that feature richness plays a crucial role in preventing mode collapse. Nevertheless, despite mitigating mode collapse, this initialization strategy achieves an FID of 6.01 when the teacher model is pre-trained on all 90 non-overlapping classes in CIFAR-100, which is significantly worse than the FID (2.39) obtained when directly using CIFAR-10 as the pre-training dataset. This suggests that while feature richness is essential for stabilizing training, functional mapping initialization remains important for achieving higher sample quality.

\vspace{-1ex}
\section{Conclusions}\label{sec:conclusion}
\vspace{-1ex}
This paper introduced \textbf{\textsc{DiffRatio}}, a new approach for training one-step diffusion models by directly estimating the score difference between the one-step model and the true data distribution across diffusion time steps with a lightweight density ratio network. Our approach mitigates the gradient estimation bias in score-based distillation and reduces the size of auxiliary networks. Furthermore, the learned density ratio network can be leveraged to unlock a principled inference-time parallel scaling without external rewards or additional sequential computation. DiffRatio achieved competitive generation quality on image synthesis benchmarks, outperforming most approaches with full teacher supervision.
Limitations and future work are discussed in Appendix~\ref{appendix:limitations}.

\clearpage

\begin{ack}
We would like to thank Prof. Mingyuan Zhou for the useful discussion.
MZ and DB acknowledge funding from AI Hub in Generative Models, under grant EP/Y028805/1.
DB is also supported by the Engineering and Physical Sciences Research Council (grant number UKRI5028).
JH is supported by the University of Cambridge Harding Distinguished Postgraduate Scholars Programme. 
ZO is supported by the Lee Family Scholarship.  
JMHL acknowledges support from a Turing AI Fellowship under grant EP/V023756/1. 
Part of the work was done when WC was a PhD student at University of Cambridge and Max Planck Institute for Intelligent Systems.
\end{ack}

\bibliography{neurips_2026}
\bibliographystyle{plain}

\clearpage
\appendix

\section{Diffusive Kullback–Leibler (DiKL) Divergence}\label{appendix:dikl}

\subsection{Validity of DiKL Divergence}\label{appendix:dikl_valid}

We follow the derivation from Spread Divergence~\cite{zhang2020spread} and provide a simple proof of the validity of the diffusive KL (DiKL) divergence. See \cite{zhang2020spread} for a generalized proof that includes cases with non-absolutely continuous distributions and non-Gaussian kernels.

Recall that DiKL is defined as
\begin{align}
    \mathrm{DiKL}_{\mathcal{K}}(q_\theta(x_0) || p_d(x_0)) = \sum_{t=1}^T w(t) \, \mathrm{KL}(q_\theta(x_t) || p_d(x_t)),
\end{align}
where \( w(t) > 0 \), and the noisy model density \( q_\theta(x_t) \) and noisy data density \( p_d(x_t) \) are respectively defined as:
\begin{align}
    q_\theta(x_t) &= \int  k_t(x_t | x_0) q_\theta(x_0) dx_0, \\
    p_d(x_t) &= \int k_t(x_t | x_0) p_d(x_0) dx_0,
\end{align}
with \( k_t(x_t | x_0) \) denoting the Gaussian transition kernel. We will show that
\begin{align}
\mathrm{DiKL}_{\mathcal{K}}(q_\theta(x_0) || p_d(x_0)) = 0 \iff q_\theta(x_0) = p_d(x_0).
\end{align}
Since \( w(t) > 0 \) and the KL divergence is non-negative, it suffices to show that:
\begin{align}
\mathrm{KL}(q_\theta(x_t) || p_d(x_t)) = 0  \iff q_\theta(x_0) = p_d(x_0).
\end{align}
First, by definition of KL divergence, we have 
\begin{align} 
\mathrm{KL}(q_\theta(x_t) || p_d(x_t)) = 0 \iff q_\theta(x_t) = p_d(x_t). 
\end{align}
Now, assume that \( k_t(\epsilon) = \mathcal{N}(0, \sigma^2 I) \), and rewrite the noisy densities as convolutions:
\begin{align}
    q_\theta(x_t) &= (q_\theta * k_t)(x_t), \\
    p_d(x_t) &= (p_d * k_t)(x_t).
\end{align}
Suppose \( q_\theta(x_t) = p_d(x_t) \). Applying the Fourier transform \( \mathcal{F} \), we obtain:
\begin{align}
\mathcal{F}(q_\theta * k_t) &= \mathcal{F}(q_\theta) \cdot \mathcal{F}(k_t), \\
\mathcal{F}(p_d * k_t) &= \mathcal{F}(p_d) \cdot \mathcal{F}(k_t).
\end{align}
Since the characteristic function of Gaussian $\mathcal{F}(k_t)>0$, we have:
\begin{align}
q_\theta(x_t) = p_d(x_t) &\iff
\mathcal{F}(q_\theta) \cdot \cancel{\mathcal{F}(k_t)} = \mathcal{F}(p_d) \cdot \cancel{\mathcal{F}(k_t)} \\
&\iff  \mathcal{F}(q_\theta)=\mathcal{F}(p_d) \\
&\iff q_\theta(x_0)=p_d(x_0).
\end{align}
This competes the proof.

\subsection{Derivation of the Density-Ratio Gradient Estimator for DiKL Divergence}\label{appendix:dikl_derivation}

Without loss of generality, we first consider DiKL with a single kernel $k_t$.
We will first follow \cite{he2024training} and show that the gradient of DiKL w.r.t. the model parameter $\theta$ is given by
\begin{equation}
    \nabla_\theta \mathrm{DiKL}_{k_t}(q_\theta || p_d) 
    =\int q_\theta(xt) \left( \nabla_{x_t} \log q_\theta(x_t) - \nabla_{x_t} \log p_d(x_t) \right) \frac{\partial x_t}{\partial \theta} dx_t.
\end{equation}
Recall that the DiKL divergence for a single kernel $k_t$ is defined as
\begin{equation}
    \mathrm{DiKL}_{k_t}(q_\theta || p_d) 
    = \int  \left( \log q_\theta(x_t) - \log p_d(x_t) \right) q_\theta(x_t) dx_t.
\end{equation}
We first reparameterize $x_t$ as a function of $z$ and $\epsilon$:
\begin{equation}
    x_t = \alpha_t g_{\theta}(z) + \sigma_t \epsilon_t\equiv h_{\theta}(z,\epsilon_t),
\end{equation}
where $z\sim p(z)\equiv \mathcal{N}(z|0,I)$ and $\epsilon_t\sim p(\epsilon_t)\equiv\mathcal{N}(\epsilon_t|0,I)$. It then follows that
{\footnotesize
\begin{align}
    \nabla_\theta \mathrm{DiKL}_{k_t}(q_\theta(x_0) || p_d(x_0))
    &= \nabla_\theta \int \left( \log q_\theta(x_t) - \log p_d(x_t) \right) q_\theta(x_t) dx_t \\
    &= \nabla_\theta \iiint \left( \log q_\theta(x_t) - \log p_d(x_t) \right)\delta(x_t-h_{\theta}(z,\epsilon_t)) p(z) p(\epsilon_t) dx_tdzd\epsilon\\
    &= \nabla_\theta \iint \left( \log q_\theta(x_t) - \log p_d(x_t) \right)|_{x_t=h_{\theta}(z,\epsilon_t)} p(z) p(\epsilon_t) dzd\epsilon\\
    &= \int \left(\nabla_\theta \log q_\theta(x_t)+\nabla_{x_t} \log q_\theta(x_t)\frac{\partial x_t}{\partial \theta} - \nabla_{x_t} \log p_d(x_t)\frac{\partial x_t}{\partial \theta} \right) q_{\theta}(x_t)dx_t \\
    &= \int \left(\nabla_{x_t} \log q_\theta(x_t)\frac{\partial x_t}{\partial \theta} - \nabla_{x_t} \log p_d(x_t)\frac{\partial x_t}{\partial \theta} \right) q_{\theta}(x_t)dx_t,
\end{align}
}
where the last line follows since
\begin{equation}
    \int \nabla_{\theta}\log q_\theta(x_t)q_\theta(x_t) dx_t = \int \nabla_{\theta} \, q_\theta(x_t)dx_t =\nabla_{\theta}\int  q_\theta(x_t)dx_t = \nabla_{\theta} 1=0.
\end{equation}
For a set of kernels $\mathcal{K}=\{k_1,\cdots,k_T\}$, it then follows that
\begin{align}
    \nabla_\theta \mathrm{DiKL}_{\mathcal{K}}(q_\theta || p_d) 
    &=\sum_{t=1}^Tw(t)\int q_\theta(x_t) \left( \nabla_{x_t} \log q_\theta(x_t) - \nabla_{x_t} \log p_d(x_t) \right) \frac{\partial x_t}{\partial \theta} dx_t \\
    &= \sum_{t=1}^Tw(t)\int q_\theta(x_t) \left( \nabla_{x_t} \log \frac{q_\theta(x_t)}{p_d(x_t)} \right) \frac{\partial x_t}{\partial \theta} dx_t \\
    &= \sum_{t=1}^Tw(t)\int q_\theta(x_t)  \nabla_{x_t} \textrm{ logit}(1-c_*(x_t,t))  \frac{\partial x_t}{\partial \theta} dx_t,
\end{align}
where the last line follows by Eq.~\ref{eq:class_ratio_est_dikl}. 
This completes the proof.

\section{Diffusive Jensen-Shannon (DiJS) Divergence}\label{appendix:dijs}

\subsection{Validity of DiJS Divergence}
We follow a similar derivation for the validity of DiKL in Appendix~\ref{appendix:dikl_valid}.
Recall that DiJS is defined as
\begin{align}
    \mathrm{DiJS}_{\mathcal{K}}(q_\theta(x_0) || p_d(x_0)) = \sum_{t=1}^T w(t) \, \mathrm{JS}(q_\theta(x_t) || p_d(x_t)),
\end{align}
where \( w(t) > 0 \), and the noisy model density \( q_\theta(x_t) \) and noisy data density \( p_d(x_t) \) are respectively defined as:
\begin{align}
    q_\theta(x_t) &= \int  k_t(x_t | x_0) q_\theta(x_0) dx_0, \\
    p_d(x_t) &= \int k_t(x_t | x_0) p_d(x_0) dx_0,
\end{align}
with \( k_t(x_t | x_0) \) denoting the Gaussian transition kernel. We will show that
\begin{align}
\mathrm{DiJS}_{\mathcal{K}}(q_\theta(x_0) || p_d(x_0)) = 0 \iff q_\theta(x_0) = p_d(x_0).
\end{align}
Since \( w(t) > 0 \) and the JS divergence is non-negative, it suffices to show that:
\begin{align}
\mathrm{JS}(q_\theta(x_t) || p_d(x_t)) = 0  \iff q_\theta(x_0) = p_d(x_0).
\end{align}
First, by definition of JS divergence, we have 
\begin{align} 
\mathrm{JS}(q_\theta(x_t) || p_d(x_t)) = 0 \iff q_\theta(x_t) = p_d(x_t). 
\end{align}
Now, assume that \( k_t(\epsilon) = \mathcal{N}(0, \sigma^2 I) \), and rewrite the noisy densities as convolutions:
\begin{align}
    q_\theta(x_t) &= (q_\theta * k_t)(x_t), \\
    p_d(x_t) &= (p_d * k_t)(x_t).
\end{align}
Suppose \( q_\theta(x_t) = p_d(x_t) \). Applying the Fourier transform \( \mathcal{F} \), we obtain:
\begin{align}
\mathcal{F}(q_\theta * k_t) &= \mathcal{F}(q_\theta) \cdot \mathcal{F}(k_t), \\
\mathcal{F}(p_d * k_t) &= \mathcal{F}(p_d) \cdot \mathcal{F}(k_t).
\end{align}
Since the characteristic function of Gaussian $\mathcal{F}(k_t)>0$, we have:
\begin{align}
q_\theta(x_t) = p_d(x_t) &\iff
\mathcal{F}(q_\theta) \cdot \cancel{\mathcal{F}(k_t)} = \mathcal{F}(p_d) \cdot \cancel{\mathcal{F}(k_t)} \\
&\iff  \mathcal{F}(q_\theta)=\mathcal{F}(p_d) \\
&\iff q_\theta(x_0)=p_d(x_0).
\end{align}
This competes the proof.

\subsection{Derivation of the Density-Ratio Gradient Estimator for DiJS Divergence}
\label{appendix:dijs_derivation}

Recall that the Jensen--Shannon (JS) divergence between two densities $q$ and $p$ is
\begin{align}
    \mathrm{JS}(q \,\|\, p) 
    = \frac{1}{2}\mathrm{KL}\!\left(q \,\Big\|\, \frac{q+p}{2}\right)
    + \frac{1}{2}\mathrm{KL}\!\left(p \,\Big\|\, \frac{q+p}{2}\right).
\end{align}
Define the mixture density at diffusion time $t$ as
\begin{align}
    m(x_t) 
    \equiv\frac{q_\theta(x_t)+p_d(x_t)}{2}.
\end{align}
We first consider a single diffusion kernel $k_t$ and then extend the result to the weighted multi-kernel setting.

\paragraph{Density-ratio identity via an optimal classifier.}
Following the GAN view~\cite{goodfellow2014generative}, consider a time-conditional binary classification problem
between data and model samples at diffusion time $t$, where $y=1 \sim p_d(x_t)$ and $ y=0 \sim q_\theta(x_t)$,
let the Bayes-optimal classifier be
$c_*(x_t,t) \equiv p(y=1\mid x_t,t).$
Then
\begin{align}
    c_*(x_t,t)=\frac{p_d(x_t)}{p_d(x_t)+q_\theta(x_t)},
    \qquad
    1-c_*(x_t,t)=\frac{q_\theta(x_t)}{p_d(x_t)+q_\theta(x_t)}.
\end{align}
Consequently, the density ratio appearing in $\mathrm{KL}(q_\theta\|m)$ can be written as
\begin{align}
    \frac{q_\theta(x_t)}{m(x_t)}
    \;=\;
    \frac{q_\theta(x_t)}{\frac{1}{2}(p_d(x_t)+q_\theta(x_t))}
    \;=\;
    2\frac{q_\theta(x_t)}{p_d(x_t)+q_\theta(x_t)}
    \;=\;
    2\bigl(1-c_*(x_t,t)\bigr).
    \label{eq:q_over_m_classifier}
\end{align}

\paragraph{Gradient of the DiJS objective under the GAN approximation.}
Consider the first JS term:
\begin{align}
    \frac{1}{2}\mathrm{KL}\!\left(q_\theta(x_t)\,\|\,m(x_t)\right)
    =
    \frac{1}{2}\int q_\theta(x_t)\log\!\left(\frac{q_\theta(x_t)}{m(x_t)}\right)\,dx_t.
\end{align}
Under the standard GAN approximation, we treat the density-ratio estimate (equivalently the optimal classifier)
as fixed when differentiating w.r.t.\ $\theta$, i.e., we detach $m(x_t)$.\footnote{This is justified by the envelope
theorem~\cite{milgrom2002envelope} for the inner maximization over the classifier parameters, and corresponds to a two-timescale training regime
where the classifier tracks its optimum for the current $\theta$. See also Theorem 2.4 in \cite{arjovsky2017towards} for a discussion of vanishing gradients in the GAN setting.}
Using the same reparameterization argument as in Appendix~\ref{appendix:dikl_derivation}, we obtain
\begin{align}
    \nabla_\theta \frac{1}{2}\mathrm{KL}\!\left(q_\theta(x_t)\,\|\,m(x_t)\right)
    &\approx
    \frac{1}{2}\int q_\theta(x_t)\,
    \nabla_{x_t}\log\!\left(\frac{q_\theta(x_t)}{m(x_t)}\right)\,
    \frac{\partial x_t}{\partial \theta}\,dx_t.
\end{align}
Substituting \eqref{eq:q_over_m_classifier} and dropping the constant $\log 2$ (whose gradient w.r.t.\ $x_t$ is zero),
this becomes
\begin{align}
    \nabla_\theta \frac{1}{2}\mathrm{KL}\!\left(q_\theta(x_t)\,\|\,m(x_t)\right)
    &\approx
    \frac{1}{2}\int q_\theta(x_t)\,
    \nabla_{x_t}\log\bigl(1-c_*(x_t,t)\bigr)\,
    \frac{\partial x_t}{\partial \theta}\,dx_t.
    \label{eq:single_t_dijs_grad}
\end{align}
For the second JS term, $\frac{1}{2}\mathrm{KL}(p_d\|m)$, the data density $p_d$ does not depend on $\theta$, and
under the same detach-$m$ approximation its contribution to $\nabla_\theta$ is ignored. Therefore, the resulting
gradient estimator is given by \eqref{eq:single_t_dijs_grad}.

\section{Discussion on the Consistency of the Diffusive Density-Ratio Estimator}
\label{appendix:convergence}

Appendices~\ref{appendix:dikl} and \ref{appendix:dijs} establish that
DiKL and DiJS are valid divergences: a zero value of either divergence
implies equality of the original model and data distributions.
This identifies the desired population optimum. To justify the
classifier-based score-difference estimates used in our gradient
formulas, however, we must distinguish consistency of the estimated
log-density-ratio values from consistency of their input gradients.
The latter requires an additional regularity condition, which we state
and prove below.

\paragraph{Consistency of the log-density-ratio values.}
Throughout this discussion, we will fix the generator parameters $\theta$ and
a noise level $t>0$. We train a time-conditional logistic classifier
$c_\eta(x_t,t)$ to distinguish real samples from generated samples, using equal class priors. The population-optimal classifier is
\begin{align}
c^\star(x_t,t)
=
\frac{p_d(x_t)}{p_d(x_t)+q_{\theta}(x_t)}.
\end{align}
Consequently, its log odds recover the log-density ratio:
\begin{align}
\ell(x_t,t)
:=
\log\frac{q_{\theta}(x_t)}{p_d(x_t)}
=
\log\frac{1-c^\star(x_t,t)}{c^\star(x_t,t)}.
\end{align}
Let $\{\hat{\eta}_n\}_{n\ge 1}$ be a sequence of classifier parameter
estimates, where $n$ indexes the estimators and may represent classifier
training iterations with the generator parameters $\theta$ held fixed.
Define
\begin{align}
\widehat{\ell}_n(x_t,t)
:=
\log\frac{1-c_{\hat{\eta}_n}(x_t,t)}
              {c_{\hat{\eta}_n}(x_t,t)}.
\end{align}

Under suitable statistical and optimization conditions, a well-specified
logistic classifier trained by maximum likelihood consistently estimates
the class probabilities and hence the log-density ratio; see the related
NCE consistency result in Theorem~2 of \cite{gutmann2010noise}.
However, consistency of the estimated values alone does not imply
consistency of their input gradients. The following theorem gives
a sufficient additional smoothness condition for gradient consistency.

\begin{theorem}[Gradient consistency of the diffusive density-ratio estimator]
\label{thm:diffusive_ratio_gradient_consistency}
Fix $\theta$ and $t>0$, and let $\ell$ and $\widehat{\ell}_n$ be the log-density ratio and its estimate defined
as above. Suppose that $\ell(\cdot,t)$ is continuously differentiable
and that the following conditions hold:
\begin{enumerate}
    \item \textbf{Pointwise log-density-ratio consistency.}
    For every $x$, we have
    \begin{align}
    \widehat{\ell}_n(x,t)
    \xrightarrow{\mathbb{P}}
    \ell(x,t)
    \quad\text{as }n\to\infty.
    \end{align}
    \item \textbf{Uniform local regularity of the estimated gradients.}
    Each $\widehat{\ell}_n(\cdot,t)$ is continuously differentiable
    almost surely. For every compact set $K$, there
    exists a deterministic constant $L_K<\infty$, independent of $n$,
    such that, almost surely,
    \begin{align}
    \left\|
    \nabla_x\widehat{\ell}_n(x,t)
    -
    \nabla_x\widehat{\ell}_n(y,t)
    \right\|
    \le L_K\|x-y\|,
    \quad x,y\in K,\quad n\ge 1.
    \end{align}
\end{enumerate}
Then, for every $x$, the input gradient of the estimated log-density ratio is a
pointwise consistent estimator of the score difference:
\begin{align}
\nabla_x\widehat{\ell}_n(x,t)
\xrightarrow{\mathbb{P}}
\nabla_x\ell(x,t)
=
\nabla_x\log q_{\theta,t}(x)
-
\nabla_x\log p_t(x).
\end{align}
\end{theorem}

\begin{proof}
For notation simplicity, we omit $t$ in the notation. Fix $x$ and a coordinate
unit vector $e_j$. Choose a compact ball $K$ containing $x$ in its
interior, and take $h>0$ sufficiently small that
$x+s e_j\in K$ for all $s\in[0,h]$. Define the finite difference along direction $e_j$:
\begin{align}
D_{h,j}f(x)
:=
\frac{f(x+h e_j)-f(x)}{h}.
\end{align}
Let $\nabla_j$ be the directional derivative along $e_j$. By the fundamental theorem of calculus, we have
\begin{align}
D_{h,j}\widehat{\ell}_n(x)
-
\nabla_j\widehat{\ell}_n(x)
=
\frac{1}{h}
\int_0^h
\left(
\nabla\widehat{\ell}_n(x+s e_j)
-
\nabla\widehat{\ell}_n(x)
\right)^\top e_j\,ds.
\end{align}
Since $\|e_j\|=1$, the uniform local Lipschitz condition implies almost surely we have
\begin{align}
\left\lVert
D_{h,j}\widehat{\ell}_n(x)
-
\nabla_j\widehat{\ell}_n(x)
\right\rVert
\le
\frac{1}{h}\int_0^h L_K s\,ds
=
\frac{L_Kh}{2}
\end{align}
Applying the triangle inequality therefore gives
\begin{align}
\left\lVert
\nabla_j\widehat{\ell}_n(x)-\nabla_j\ell(x)
\right\rVert
&\le
\left\lVert
\nabla_j\widehat{\ell}_n(x)-D_{h,j}\widehat{\ell}_n(x)
\right\rVert
+\left\lVert
D_{h,j}\widehat{\ell}_n(x)-\nabla_j\ell(x)
\right\rVert\\
&\le
\frac{L_Kh}{2}
+\left\lVert
D_{h,j}\widehat{\ell}_n(x)-D_{h,j}\ell(x)
\right\rVert
+
\left\lVert
D_{h,j}\ell(x)-\nabla_j\ell(x)
\right\rVert\\
&\le
\frac{L_Kh}{2}
+
\frac{
\left|\widehat{\ell}_n(x+h e_j)-\ell(x+h e_j)\right|
+
\left|\widehat{\ell}_n(x)-\ell(x)\right|
}{h}\nonumber
\\
&\quad+
\left\lVert
D_{h,j}\ell(x)-\nabla_j\ell(x)
\right\rVert.
\end{align}

Let $\varepsilon>0$. Because $\ell$ is differentiable and $L_K$ is
independent of $n$, we can first choose a fixed $h>0$ sufficiently
small that the sum of the first and last terms is less than
$\varepsilon/2$. For this fixed $h$, the middle term converges to zero
in probability by pointwise consistency at $x$ and $x+h e_j$. Hence,
\begin{align}
\mathbb{P}\!\left(
\left|
\nabla_j\widehat{\ell}_n(x)-\nabla_j\ell(x)
\right|>\varepsilon
\right)
\longrightarrow 0.
\end{align}
Applying this argument to each of the finitely many coordinates
$j=1,\ldots,d$ proves
\begin{align}
\nabla\widehat{\ell}_n(x)
\xrightarrow{\mathbb{P}}
\nabla\ell(x).
\end{align}
This completes the proof.
\end{proof}

\paragraph{Role of Gaussian diffusion and scope of the result.}
Support overlap is a potential obstacle to density-ratio estimation
for implicit generators: the original distributions $q_\theta$ and
$p_d$ may be singular or have disjoint supports. Gaussian diffusion
removes this obstacle at every fixed noise level with strictly positive
noise variance. Specifically, for a forward kernel
$x_t=a_t x_0+\sigma_t\epsilon$, where
$\epsilon\sim\mathcal{N}(0,I)$ and $\sigma_t>0$, the noised densities are
\begin{align}
p_t(x)
=
\int \varphi_{\sigma_t}(x-a_tu)\,p_d(du),
\qquad
q_{\theta,t}(x)
=
\int \varphi_{\sigma_t}(x-a_tu)\,q_\theta(du),
\end{align}
where $\varphi_{\sigma_t}$ is the density of
$\mathcal{N}(0,\sigma_t^2I)$ and $p_d,q_\theta$ here denote the original
clean probability measures. Both noised densities are strictly positive and
smooth on $\mathbb{R}^d$, even when the original distributions do not
admit densities. Their log-density ratio is therefore smooth, and the
support condition required for density-ratio estimation is satisfied.

Thus, performing ratio estimation at $t>0$ removes the need to assume
support overlap at $t=0$. Statistical consistency still requires the
appropriate estimation assumptions, and gradient consistency
additionally requires the regularity condition in
Theorem~\ref{thm:diffusive_ratio_gradient_consistency}. In particular,
Gaussian smoothing of the target distributions does not automatically
impose this condition on the learned estimators.

\section{Detailed Setup of Toy Experiment}\label{appendix:toy_setup}

This section provides the detailed experimental setup for the gradient bias study in Section~\ref{sec:toy}.

\textbf{Data Distribution.}
The data distribution $p_d(x_0)$ is a mixture of Gaussians with 40 components in 2D. Component means are sampled uniformly in $[-40, 40]^2$ and variances are sampled log-uniformly, providing a multi-modal distribution with varying local densities. This setup allows closed-form computation of the true score $\nabla_{x_t} \log p_d(x_t)$ at any noise level $t$.

\textbf{One-Step Diffusion Model.}
The one-step diffusion model $q_\theta(x_0)$ is an implicit generative model parameterized by a 4-layer MLP with hidden dimension 400 and SiLU activations, conditioned on the noise level $t$. The model has been pre-trained to approximate $p_d$. Since $q_\theta$ lacks a closed-form density, we estimate its score using kernel density estimation (KDE) with 10,000 generated samples and bandwidth selected via Scott's rule. This provides the ground-truth score difference:
\begin{equation}
\Delta s_t^* \equiv \Delta s^*(x_t, t) = \nabla_{x_t} \log q_\theta(x_t) - \nabla_{x_t} \log p_d(x_t).
\end{equation}

\textbf{Methods Compared.} We compare the bias of the two gradient estimators, VSD and DiffRatio. 
\begin{enumerate}
    \item \textbf{DiffRatio (ours)}: A time-conditioned classifier $c_\eta(x_t, t)$ is trained to distinguish between samples from $q_\theta(x_t)$ and $p_d(x_t)$. The classifier is a 4-layer MLP with hidden dimension 400 and SiLU activations, taking concatenated $(x_t, t)$ as input and outputting a scalar logit. The score difference is obtained via Eq.~\ref{eq:class_ratio_est_dikl}: $\widehat{\Delta s_t}=\nabla_{x_t} \operatorname{logit}(1 - c_\eta(x_t, t))$.
    \item \textbf{VSD}: Two separate score networks $s_{\psi_2}^{q_\theta}(x_t,t)$ and $s_{\psi_1}^{p_d}(x_t,t)$ are trained independently using DSM (Eqs.~\ref{eq:dsm_loss:pd} and \ref{eq:dsm_loss:q}), and the score difference is computed as their difference $\widehat{\Delta s_t}=s_{\psi_2}^{q_\theta}(x_t,t)-s_{\psi_1}^{p_d}(x_t,t)$. Both networks use the same architecture as the classifier.
\end{enumerate}

\textbf{Training Details.}
All networks are trained for 10,000 iterations with batch size 1024 using Adam optimizer with learning rate $10^{-4}$. Noise levels are sampled from a power schedule $t \sim t_{\min} + u^{1.5} \cdot (t_{\max} - t_{\min})$ where $u \sim \mathcal{U}(0,1)$, $t_{\min} = 0.1$, and $t_{\max} = 20.0$. The classifier is trained with binary cross-entropy loss using softplus for numerical stability.

\textbf{Evaluation.}
We evaluate each method by comparing estimated score differences against the KDE-based ground truth at 10 logarithmically-spaced noise levels in $[0.1, 20]$. At each noise level, we sample 1,000 evaluation points from the noised student distribution. We report the cosine similarity $\langle \widehat{\Delta s_t}, \Delta s_t^* \rangle / (\|\widehat{\Delta s_t}\| \|\Delta s_t^*\|)$, measuring directional alignment of the gradient estimates.

\section{Detailed Setup of Image Generation Experiments}\label{appendix:exp_setup}

This section provides the detailed experimental setup for the image generation experiments in Section~\ref{sec:experiments}.

We evaluate the performance of our proposed method by training one-step image diffusion models on two standard benchmarks: (1) unconditional generation on the CIFAR-10 ($32{\times}32$) dataset~\cite{krizhevsky2009learning}, (2) class-label-conditioned generation on the ImageNet ($64{\times}64$ and $512{\times}512$) dataset~\cite{deng2009imagenet}. 

Our one-step diffusion model adopts the same neural network architecture as the pre-trained score model. 
Our implementation builds upon the EDM and EDM2 codebases~\cite{karras2022elucidating,karras2024analyzing}.
Specifically, we employ the EDM U-Net model architecture \cite{karras2022elucidating} for CIFAR-10 and ImageNet $64{\times}64$, and use the EDM2-M U-Net model architecture \cite{karras2024analyzing} for ImageNet $512{\times}512$. We use the EDM preconditioning for the inputs. We choose not to use the EDM output preconditioning as we are training a one-step generator which outputs images that differ significantly from the input noise. 

Our density-ratio estimator \( c_\eta(x_t, t) \) is implemented using the encoder portion of the teacher's U-Net architecture with a sigmoid function attached to the end to produce a scalar output between $0$ and $1$. 
This network is approximately half the size of the full U-Net used for score estimation, leading to reduced size of auxiliary networks. The one-step generator \( g_\theta(z) \) is initialized using the weights from the pre-trained score model with the diffusion time argument $t_{\text{init}}{=}2.5$ fixed throughout training and sampling. 

We consider three training criteria in our framework (DiKL/DiJS/DiRM) on CIFAR-10, and use the best-performing criterion (DiJS) for higher-dimensional ImageNet experiments ($64{\times}64$ and $512{\times}512$). 
We use the variance-exploding (VE) noise schedule to define the diffusive divergences with $\sigma(t_{\text{min}})=0.002$ and $\sigma(t_{\text{max}})=80$.
We follow standard hyperparameter settings for training generative models on CIFAR-10 and ImageNet as detailed in \cite{karras2022elucidating,karras2024analyzing}. 
We use a batch size of 64 for CIFAR-10 and 4,096 for ImageNet.
The weighting function \( w(t) {=} \sigma_t^2 \) is used for for both datasets.
For CIFAR-10 and ImageNet $64{\times}64$, we use a fixed learning rate of $5\times10^{-5}$ for both generator and density ratio network. For ImageNet $512{\times}512$, we use a fixed learning rate of $10^{-4}$ for the density ratio network; for the generator we use a learning rate of $10^{-4}$ at the beginning and reduce the learning rate by a factor of $10$ when the FID does not improve for 500 ticks.
We adopt a single-step update strategy for the density-ratio estimator throughout our experiments, consistent with previous works~\cite{luo2024diff,zhou2024score}.
All experiments are conducted on 8 NVIDIA H100 80GB GPUs.

\clearpage
\section{Additional Image Generation Results}\label{appendix:additional_images}

\begin{figure}[H]
    \centering
\includegraphics[width=1\textwidth]{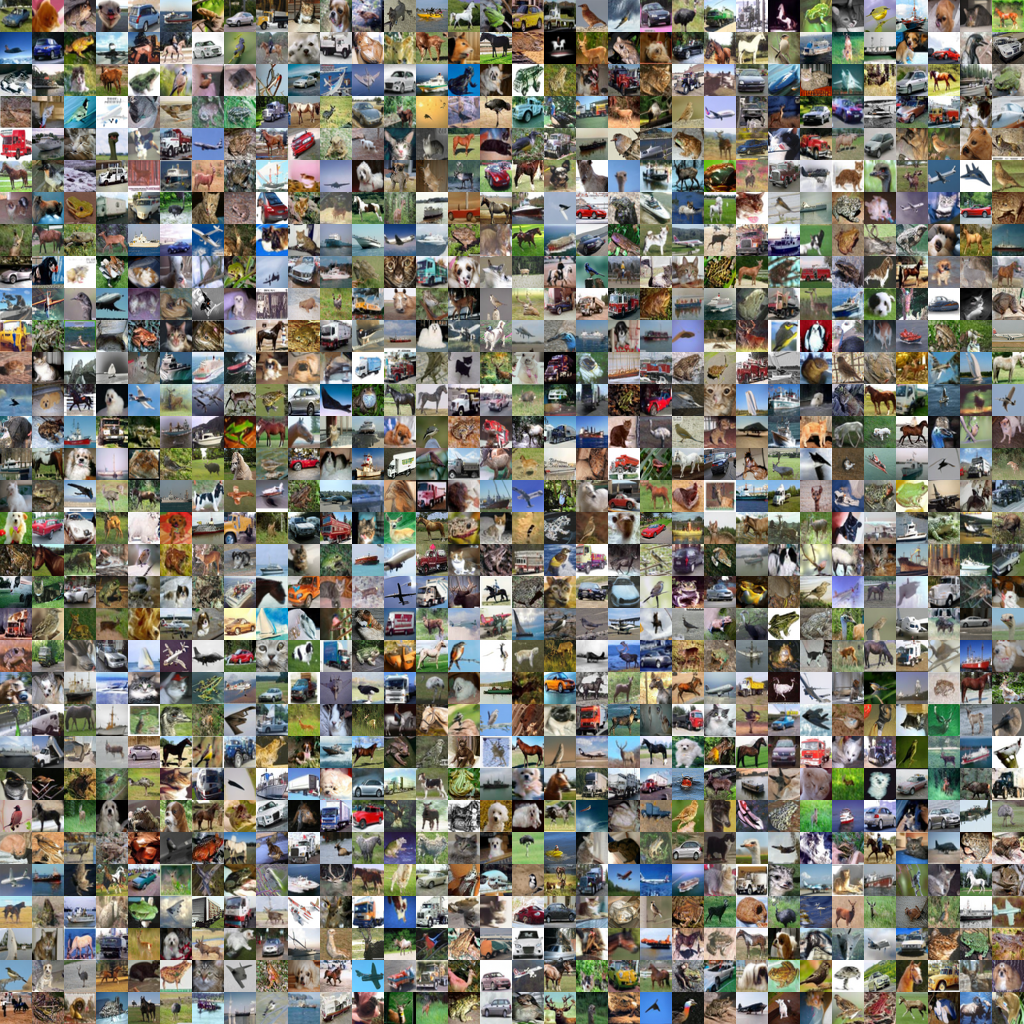}
    \caption{Visualization of samples from DiffRatio-DiJS trained on CIFAR10 (FID=2.39, IS=9.93).}
    \label{fig:dijs_cifar10_full}
\end{figure}

\begin{figure}
    \centering
    \includegraphics[width=1\textwidth]{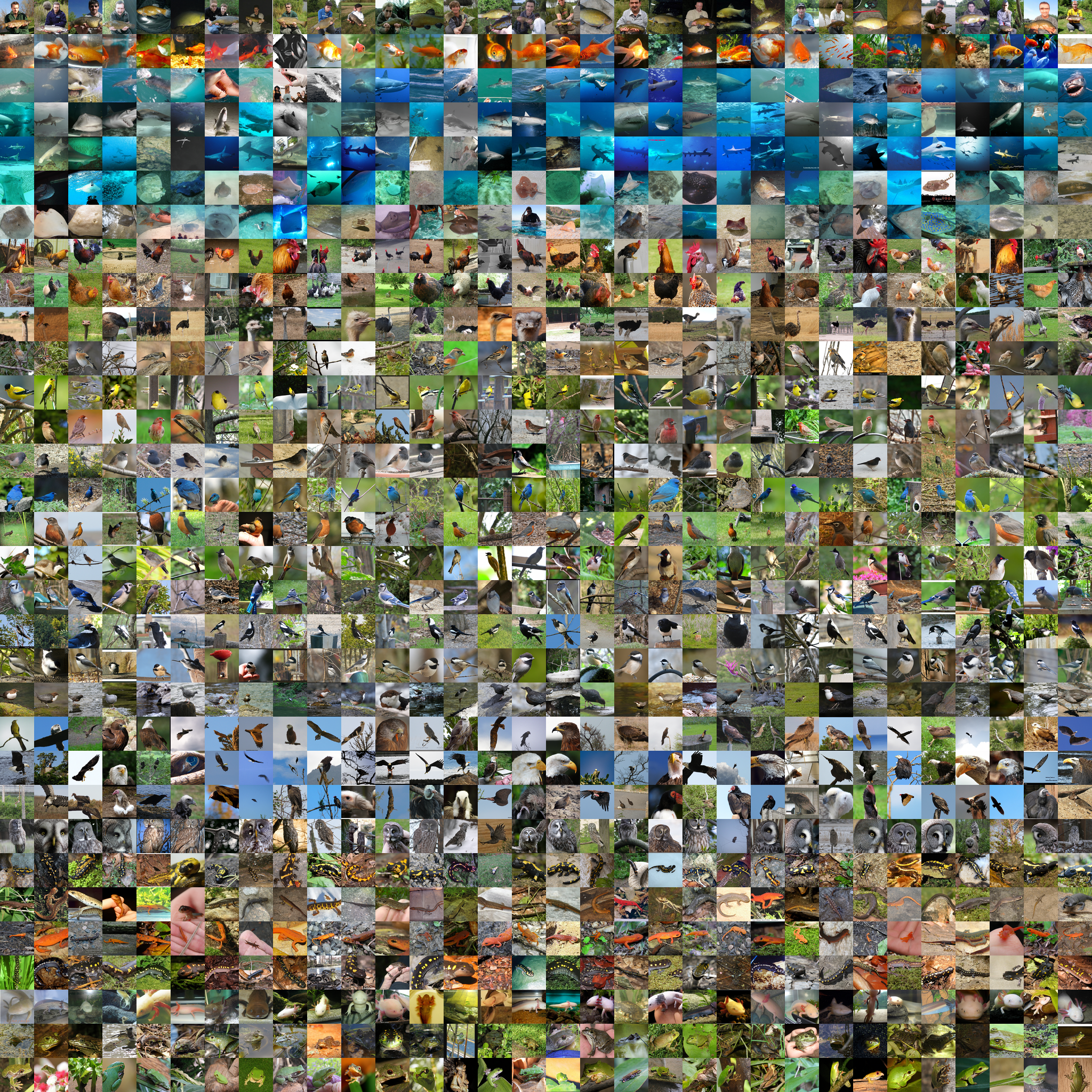}
    \caption{Visualization of samples from DiffRatio-DiJS trained on ImageNet 64${\times}$64 (FID=1.54).}
    \label{fig:dijs_imagenet64_full}
\end{figure}

\begin{figure}
    \centering
    \includegraphics[width=0.23\textwidth]{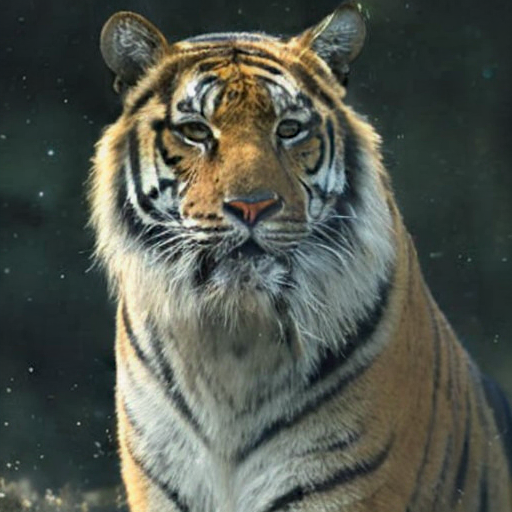}
    \includegraphics[width=0.23\textwidth]{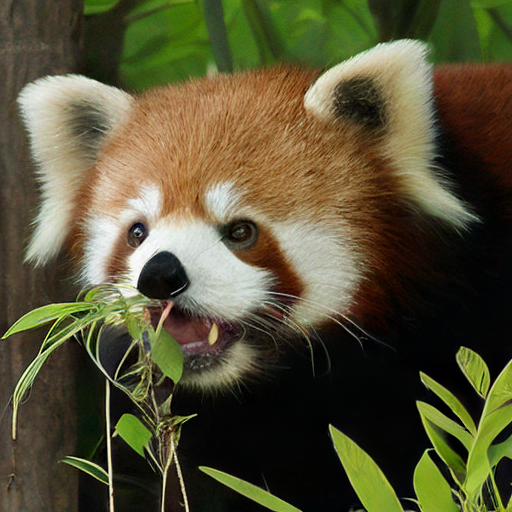}
    \includegraphics[width=0.23\textwidth]{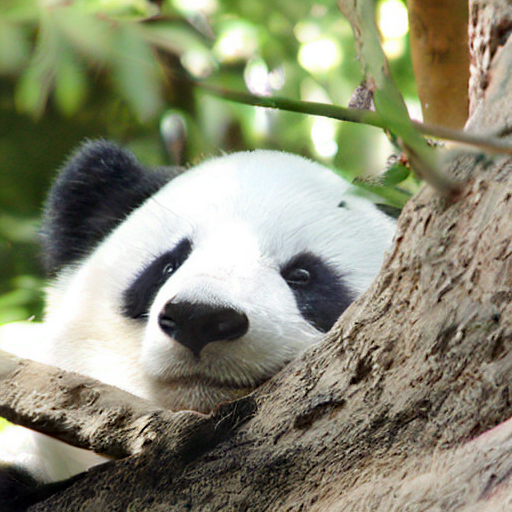}
    \includegraphics[width=0.23\linewidth]{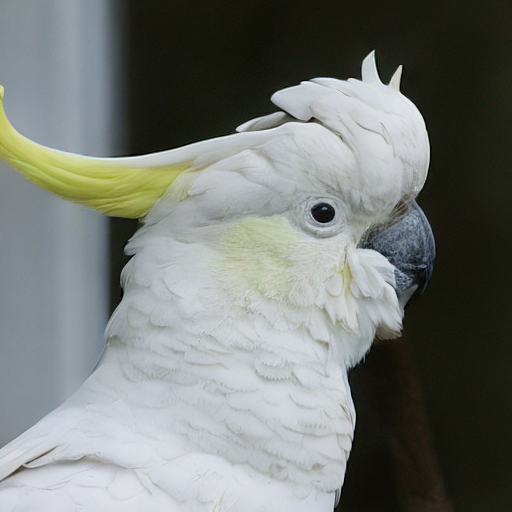}
    
    \includegraphics[width=0.23\textwidth]{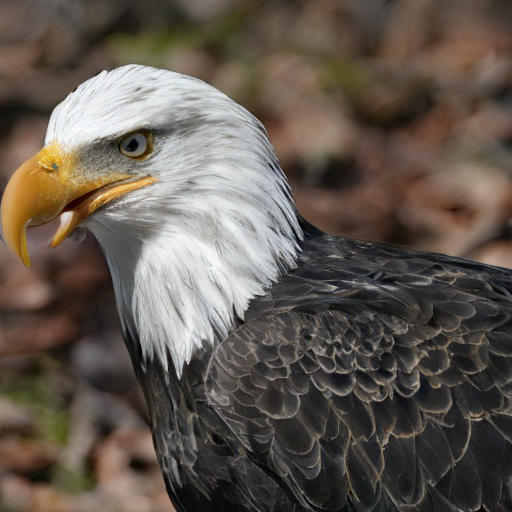}
    \includegraphics[width=0.23\textwidth]{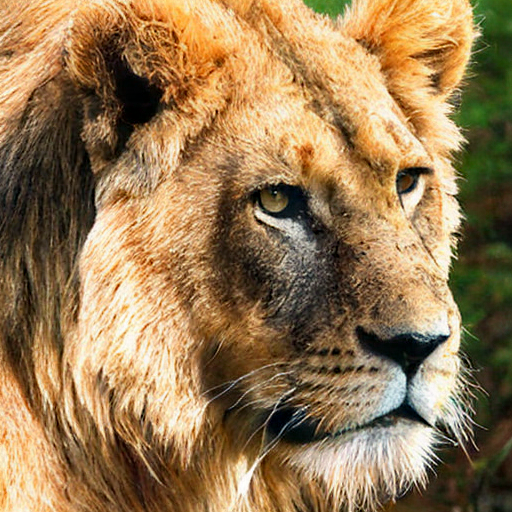}
    \includegraphics[width=0.23\textwidth]{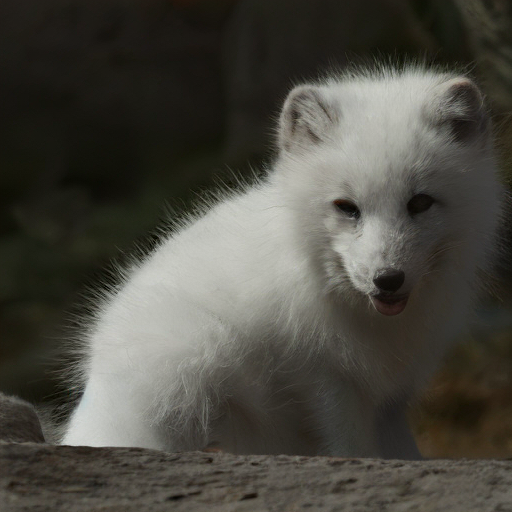}
    \includegraphics[width=0.23\linewidth]{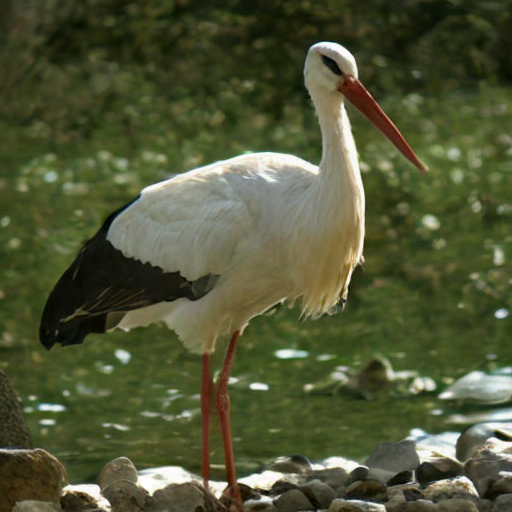}
    
    \includegraphics[width=0.23\textwidth]{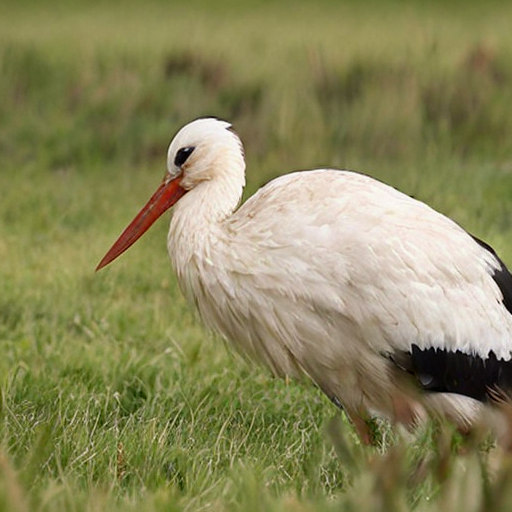}
    \includegraphics[width=0.23\textwidth]{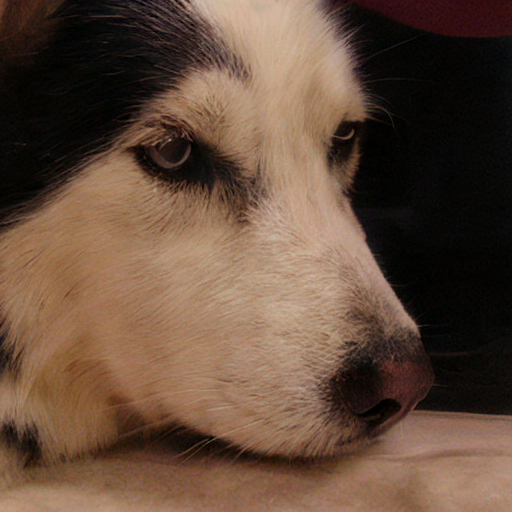}
    \includegraphics[width=0.23\textwidth]{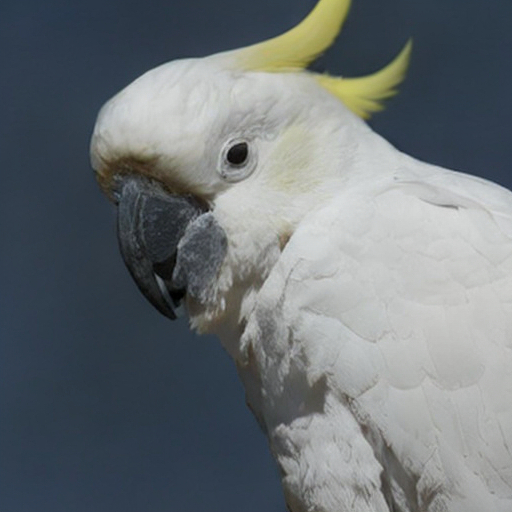}
    \includegraphics[width=0.23\linewidth]{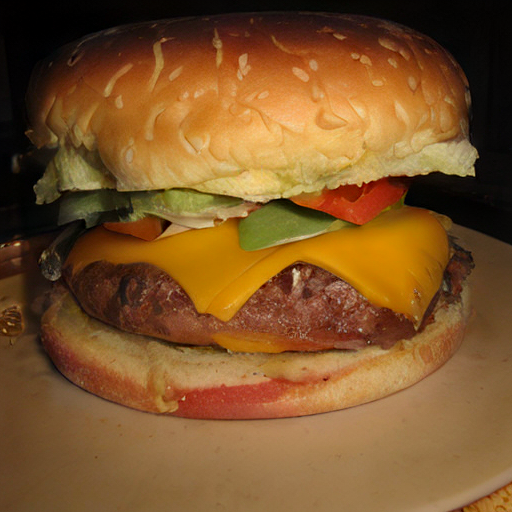}

    \includegraphics[width=0.23\textwidth]{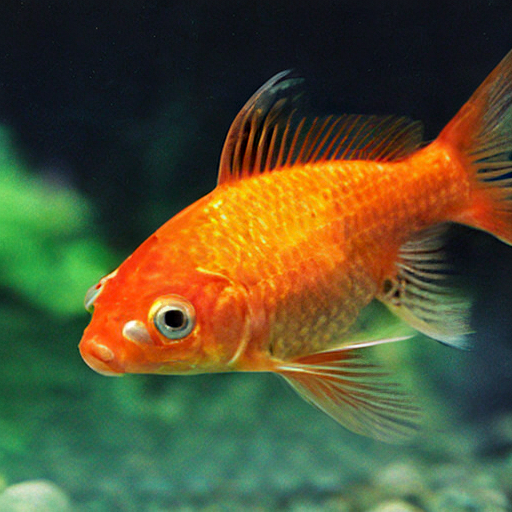}
    \includegraphics[width=0.23\textwidth]{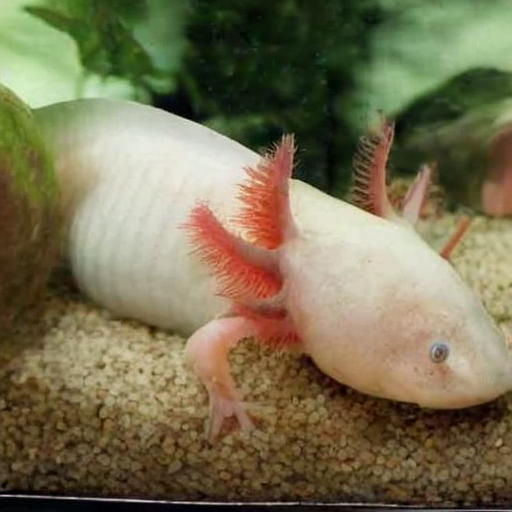}
    \includegraphics[width=0.23\textwidth]{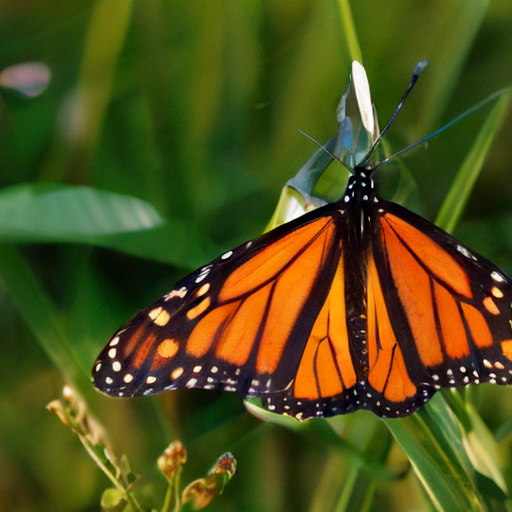}
    \includegraphics[width=0.23\linewidth]{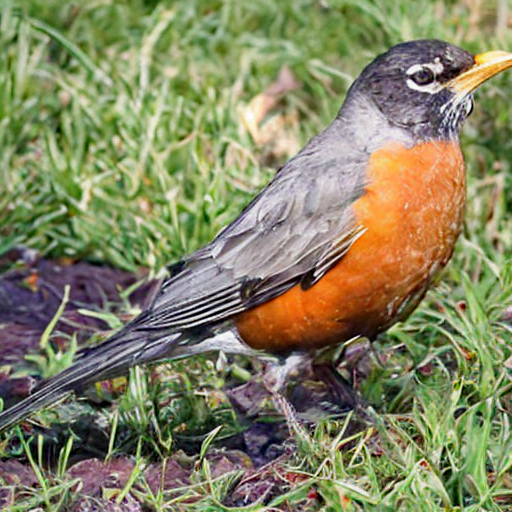}

    \includegraphics[width=0.23\textwidth]{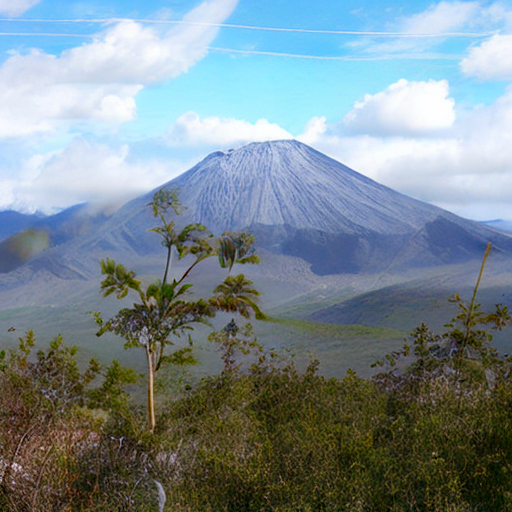}
    \includegraphics[width=0.23\textwidth]{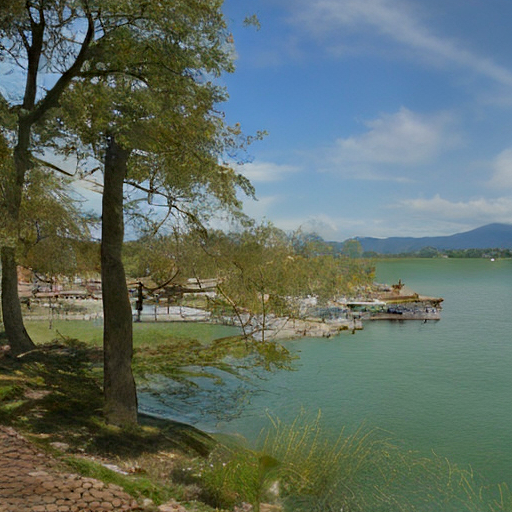}
    \includegraphics[width=0.23\textwidth]{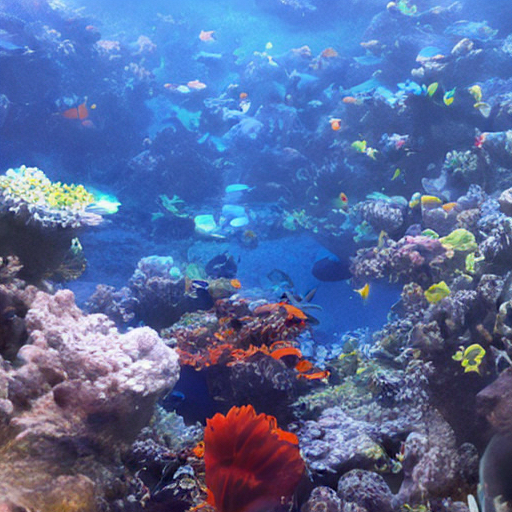}
    \includegraphics[width=0.23\textwidth]{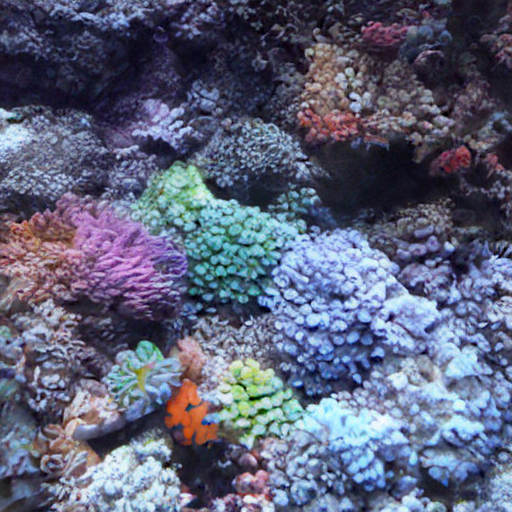}
    
    \caption{Visualization of samples from DiffRatio-DiJS-M trained on ImageNet 512${\times}$512 (FID=1.41).}
    \label{fig:dijs_imagenet512_full}
\end{figure}

\clearpage

\section{Additional Details of Variational Score Distillation}

The two-stage distillation procedure of VSD is summarized in Algorithm~\ref{alg:one_step_train}.

\begin{algorithm}[t]
    \caption{Score-based distillation for training one-step diffusion models with teacher supervision (variational score distillation, VSD)}
    \label{alg:one_step_train}
    \begin{algorithmic}[1]
        \Require Training samples $\mathcal{D} \sim p_d(x_0)$
        \Statex \textcolor{gray}{\textit{// Stage 1: Pre-train a teacher score model (requires training data)}}
        \State Pre-train the teacher model's score network $s_{\psi_1}^{p_d}(x_t,t)$ using DSM (Eq.~\ref{eq:dsm_loss:pd}) until convergence          
        \Statex \textcolor{gray}{\textit{// Stage 2: Distill a one-step student model with teacher supervision}}
        \State Initialize the one-step generator with the teacher's score network
        $g_{\theta_{\text{init}}}(\cdot)\equiv s_{\psi_1}^{p_d}(\cdot,t=t_{\text{init}})$
       \For{each training iteration}
            \State Sample time $t'$ for student score estimation
            \State Estimate the student model's score by training a score network $s_{\psi_2}^{q_\theta}(x_{t'},t')$ using DSM
         \State Update the one-step generator $g_\theta$ with $s_{\psi_2}^{q_\theta}(x_{t'},t')$ (student score) and $s_{\psi_1}^{p_d}(x_{t'},t')$ (teacher supervision) using Eq.~\ref{eq:kl_gradient}
        \EndFor
    \end{algorithmic}
\end{algorithm}

\section{Limitations and Future Work}\label{appendix:limitations}

This work takes us one step closer towards training one-step diffusion models without distillation by removing teacher supervision in the training objective. However, our proposed DiffRatio approach still requires a pre-trained score network for initializing the one-step diffusion model. A key insight from our analysis in Section \ref{sec:init} was that while teacher supervision could be removed, weight initialization from the teacher remained essential. We found that this was not due to improved latent-to-output mappings, but rather the multi-level features learned across noise levels during
diffusion training, which helped prevent mode collapse. Future work may explore unsupervised
or self-supervised pretraining to provide feature-rich initializations without requiring a pre-trained
diffusion model. 

The inference-time parallel scaling scheme proposed in this paper allows us to use the learned density ratio network at diffusion timestep $t=0$ to further improve the generation quality without external reward or extra sequential computation. However, density ratio estimation may become more accurate as the diffusion timestep $t$ increases \cite{kim2024training,na2024diffusion}. It would be interesting to investigate how to leverage larger diffusion timesteps to produce a more accurate density ratio estimation for the importance sampling procedure in the proposed inference-time parallel scaling scheme.

Also, extending our framework to other modalities such as one-step diffusion models for text-to-image and text-to-video 
remains a promising future work direction.

\end{document}